%% file: arxiv.tex
\documentclass[runningheads]{llncs}
\newcommand{\minisection}[1]{\vspace{0.0in} \noindent {\bf #1}}
\usepackage{eccv}
\usepackage{wrapfig}
\DeclareMathOperator*{\argmin}{argmin}
\usepackage{eccvabbrv}
\usepackage{multirow}
\usepackage{enumerate}
\usepackage{makecell}
\usepackage[inline]{enumitem}
\usepackage{amssymb}
\usepackage{pifont}
\newcommand{\cmark}{\ding{51}}
\newcommand{\xmark}{\ding{55}}
\usepackage{graphicx}
\usepackage{booktabs}
\usepackage[accsupp]{axessibility}  
\usepackage[pagebackref,breaklinks,colorlinks]{hyperref}
\usepackage{orcidlink}
\usepackage{xcolor,colortbl}

\begin{document}

\title{Is Retain Set All You Need in Machine Unlearning? Restoring Performance of Unlearned Models with Out-Of-Distribution Images} 

\titlerunning{SCAR}

\author{Jacopo Bonato\inst{1,*}\orcidlink{0000-0001-6751-3407} \and
Marco Cotogni\inst{1,*}\orcidlink{0000-0001-7950-7370} \and
Luigi Sabetta\inst{1,*}\orcidlink{0000-0002-0865-5891}}

\authorrunning{J.~Bonato et al.}
\institute{Leonardo Labs, Rome, Italy\\
*These authors contributed equally\\
\email{\{jacopo.bonato, marco.cotogni, luigi.sabetta\}.ext@leonardo.com}}

\maketitle

\begin{abstract}
In this paper, we introduce \textbf{S}elective-distillation for \textbf{C}lass and \textbf{A}rchitecture-agnostic unlea\textbf{R}ning (\textbf{SCAR}), a novel approximate unlearning method. SCAR efficiently eliminates specific information while preserving the model's test accuracy without using a retain set, which is a key component in state-of-the-art approximate unlearning algorithms. Our approach utilizes a modified Mahalanobis distance to guide the unlearning of the feature vectors of the instances to be forgotten, aligning them to the nearest wrong class distribution. Moreover, we propose a distillation-trick mechanism that distills the knowledge of the original model into the unlearning model with out-of-distribution images for retaining the original model's test performance without using any retain set. Importantly, we propose a self-forget version of SCAR that unlearns without having access to the forget set. We experimentally verified the effectiveness of our method, on three public datasets, comparing it with state-of-the-art methods. Our method obtains performance higher than methods that operate without the retain set and comparable w.r.t the best methods that rely on the retain set. 

\end{abstract}

\section{Introduction}
\label{sec:intro}
Over the past decades, deep learning algorithms have grown significantly in applications like image classification, detection, and segmentation. For optimal performance, these models learn from big training datasets and update their weights to capture intrinsic patterns. 
Despite their effectiveness, many ethical concerns such as potential biases affecting certain demographics \cite{Du2021_fairness} and privacy like unauthorized use of sensitive or personal data \cite{Liu2021PrivacyAS,Rigaki2023} have been raised. Additionally, methods like model inversion \cite{chen2021MI,Fredrikson2015MI,Kahla2022MI,wang2021MI,Yang2019MI,zhang2020MI,zhao2021MI}, and membership inference attacks \cite{Shokri2017, Salem2019, Song2019, Yeom2017} raised many concerns about the potential revelation of training data characteristics or can reveal particular data utilization during training. These vulnerabilities highlight the critical necessity for implementing safeguards to secure sensitive information within machine learning models. This imperative is fundamental in light of the privacy-preserving regulations that have been put into effect ( e.g. the European Union’s General Data Protection Regulation \cite{magdziarczyk2019right} and the California Consumer Privacy Act\cite{pardau2018california}).

Machine Unlearning\cite{nguyen2022survey,xu2023machine,shaik2023exploring, mercuri2022introduction} has emerged as a solution to these privacy and ethical challenges, aiming to remove specific information content from algorithms without degrading their performance. Unlearning algorithms can be classified into 3 main groups: model intrinsic, data-driven, and model agnostic. The characteristic of model-intrinsic methods is to change the deep neural network (DNN) architecture or add specific weights to induce the unlearning. The data-driven group is characterized by the usage of data to induce forgetting, and it can be roughly divided into methods that store optimization information during training (e.g. gradient updates) and methods that involve data manipulation. Similarly to this latter group, model-agnostic solutions do not require access to specific optimization-related information and operate directly on the unaltered trained model. 
Nonetheless, a common practice among these algorithms is to maintain or restore the model performance using the residual portion of the training dataset, known as retain set. Using the retain set to recalibrate models post-unlearning represents a viable strategy, however, it also introduces practical challenges. In scenarios where privacy concerns dictate, only the forget set —images to forget— might be accessible, leaving no retain set for performance restoration. This poses a significant hurdle for models designed to undergo a performance restoration phase after unlearning, potentially compromising their utility. Moreover, when large-scale datasets such as ImageNet or ImageNet-21K are involved and the forget set constitutes a minor fraction, the bulk of the data is the retain set. This disproportionality can significantly inflate the time required to restore the model to its original efficacy. Furthermore, there are even more intricate situations where also the forget set is unavailable, such as when the unlearning task entails forgetting an entire class with access only to its ID.

This paper presents a novel, model-agnostic unlearning algorithm, \textbf{S}elective-distillation for \textbf{C}lass and \textbf{A}rchitecture-agnostic unlea\textbf{R}ning (\textbf{SCAR}), which leverages metric learning and knowledge distillation \cite{hinton2015distilling} to efficiently remove targeted information and maintain model accuracy, \textbf{notably without relying on a retain set}. 
Specifically, we use the Mahalanobis distance to shift the feature vectors of the instances to forget toward the closest distribution of samples of other classes.
This distance includes important information regarding the original dataset distribution of samples resulting in an efficient metric learning-based unlearning strategy. 
Concurrently, distilling the knowledge of the original model into the unlearning model using out-of-distribution (OOD) images, allows SCAR to maintain the performance of the original model on the test set. 
The \textbf{main contributions} of this work include:  \textbf{\textit{(i)}} SCAR, a novel model-agnostic unlearning algorithm that achieves competitive unlearning performance without using retain data.
\textbf{\textit{(ii)}} A unique self-forget mechanism in class removal scenarios that operates without direct access to the forget set.
\textbf{\textit{(iii)}} Comprehensive analyses demonstrating SCAR's efficacy across different datasets and architectures in class-removal and homogeneous-removal scenarios
 \textbf{\textit{(iv)}} Experimental evidence showing SCAR's comparable or superior performance respectively to traditional unlearning methods or state-of-the-art (sota) techniques that do not utilize a retain set.
Code will be released at \url{https://github.com/jbonato1/SCAR}

\section{Related Works}
\label{sec:RW}
Machine Unlearning algorithms can be classified into three main categories: data-driven, model-intrinsic, and model-agnostic unlearning algorithms~\cite{nguyen2022survey}. 

\minisection{Data-driven.} The unlearning algorithms in this category employ a data-driven approach to remove knowledge about specific image instances. For example, methods that involve data manipulation, such as transformations or attacks, fall within this group of algorithms\cite{setlur2022adversarial,sommer2020towards}. However, this category is not limited to such models: it encompasses a broader range of approaches.  In~\cite{graves2021amnesiac,wu2020deltagrad} the authors propose similar methods for reverting the gradient updates for the instances of the classes to be removed. Despite their effectiveness, these methods require storing gradient updates for all the batches in the training set. Similarly, in \cite{neel2021descent}, the authors propose two gradient-descent-based approaches for the effective unlearning of convex models. In the first, Gaussian noise is used to perturb the gradient updates, while the second also incorporates reservoir sampling. In \cite{wu2022puma}, the authors modeled the impact of each training point on the model's training process and then compensated for the impact of removing data from the model's knowledge by re-weighting the gradient updates. In \cite{felps2020class}, the authors employed a strategy of recursively assigning an incorrect class to samples designated for forgetting until a membership inference attack no longer recognizes those samples as part of the training dataset. In \cite{bourtoule2021machine}, the authors segmented the dataset into n parts and trained n models on each segment; during inference, predictions from these models are aggregated. Only the model containing those samples is retrained when a removal request is received.
Moreover, novel retain-set-free algorithms such as Random Labels\cite{hayase2020selective} and Negative Gradient\cite{golatkar2020eternal} have been introduced. In the former approach, random labels are assigned to samples designated for forgetting to calculate the cross-entropy, while the latter employs the negative gradient obtained from these samples to optimize the loss function.
\begin{figure}[t]
    \centering
    \includegraphics[width=0.9\textwidth]{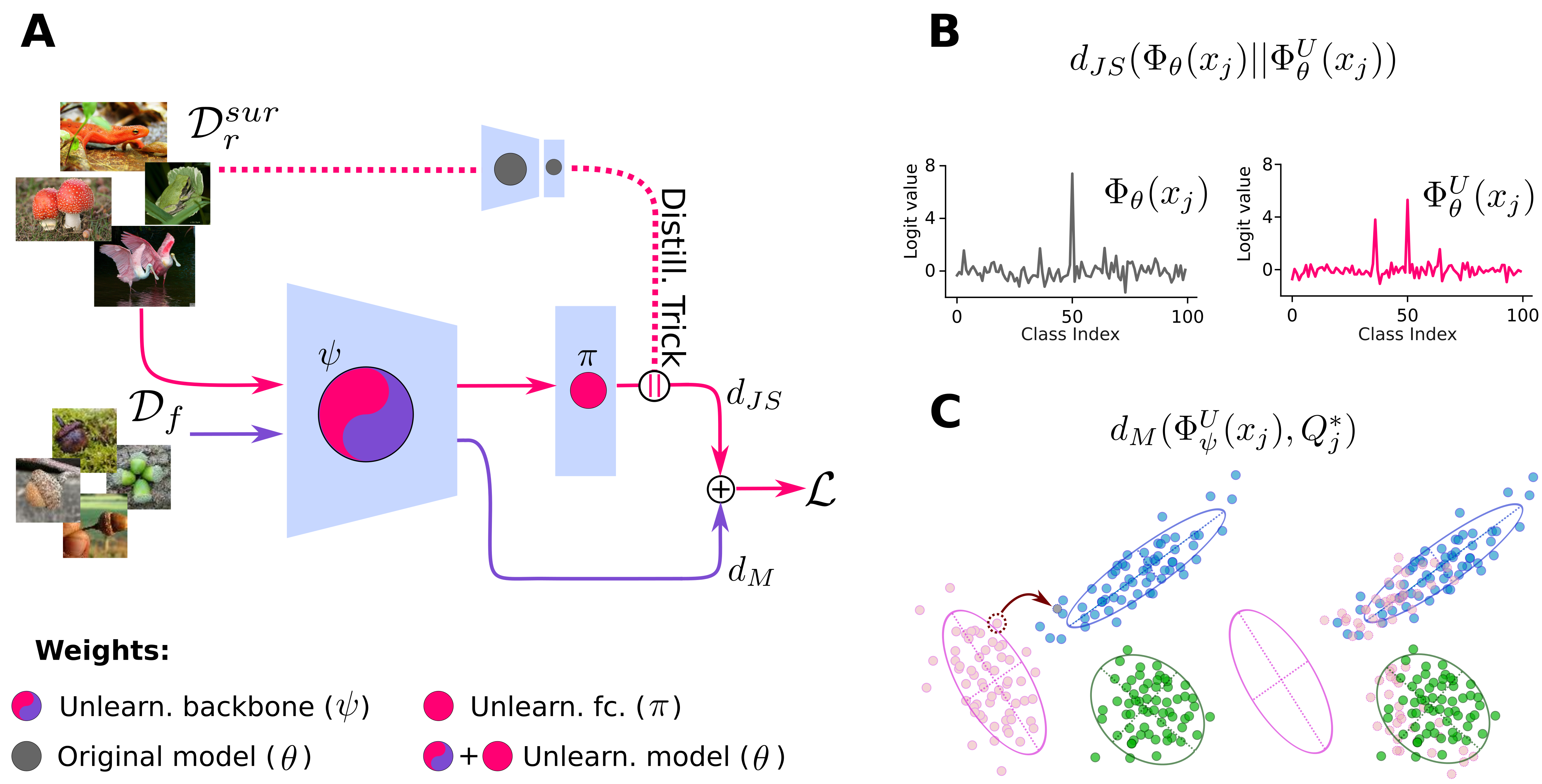}
    \vspace{-2mm}
    \caption{SCAR overview. A)SCAR scheme B)distillation-trick through J.S. divergence of the logits of $\Phi_\theta(x_j)$ and $\Phi^U_{\theta}(x_j)$. C) Representation of the feature vector distributions of samples for 3 classes before (left) and after (right) the unlearning process. Ellipses denote the $95\%$ confidence intervals of the distribution of samples.}
    \label{fig:fig1}
\end{figure}
\minisection{Model-intrinsic.} Model-intrinsic approaches to machine unlearning consist of algorithms that depend on a specific architecture. In these cases, the unlearning process is facilitated by sophisticated mechanisms tailored to function within that particular architecture. Depending on the task to be performed several architectures can be involved. For instance in \cite{golatkar2020forgetting, golatkar2020eternal} the authors proposed two different unlearning approaches for CNNs. In the former, the authors ``scrubbed'' from the weights of the model the knowledge about instances of images through the computation of Fisher's information matrix. Similarly in the latter, the authors propose a CNN-based unlearning algorithm, inspired by the neural tangent kernel, that models the weights dynamics and can remove from the information about forget data. In \cite{lin2023erm}, the authors proposed a two-phase algorithm for the approximate unlearning of CNNs. During the first phase, called ERM, the model is trained on all the data while a binary mask is learned for each layer of the CNN for each class of the dataset. Then during the second phase, called KTP, the knowledge about the classes of the forget set is removed using the previously learned masks. In \cite{pmlr-v119-guo20c}, the authors propose a certified algorithm for removing knowledge about classes on linear classifiers. This method applied a Newton step on the model's weights to significantly reduce the influence of the deleted data point, with the residual error decreasing quadratically as the training set size increases. Then, for certifying the data removal and to prevent the extraction of information from the minimal residual the training loss is perturbed. Recently, additional unlearning approaches have been developed for dealing with specific models on different learning paradigms such as Multimodal learning\cite{poppi2024removing, cheng2023multimodal}and Adversarial Learning\cite{tiwary2023adapt, Liu_2023_ICCV}.

\minisection{Model-agnostic.} This group of algorithms includes universally applicable methods across different architectures. Their characteristic is the ability to intervene in the feature/activation space to selectively erase data associated with the forget set. In \cite{kurmanji2023towards}, the authors used knowledge distillation to differentiate the logits of the forget set from those of the retain set and erase the knowledge of these examples. A similar strategy was proposed in \cite{kim2022efficient}, where distillation was combined with contrastive labeling to achieve efficient unlearning. In \cite{Chen_2023_CVPR} the authors proposed two methods for unlearning based on the kinematics of the decision boundary. The methods change the labels of the examples of the forget set to the closest wrong class and then fine-tune the model with the reassigned samples. Similarly, In \cite{cotogni2023duck} the authors proposed a metric learning-based approach that associates samples from the forget set to the closest centroid in the feature space. Through this procedure, the model preserves the feature space organization. In \cite{jia2023model} the authors proposed a model sparsification via weight pruning for unlearning data samples. In \cite{tarun2023fast}, the authors proposed to learn a matrix noise associated with the classes to be forgotten to remove their influence on the model's weights. After that, a repair step is used to recover knowledge about retain samples. 

Our approach stands out as a model-agnostic approximate unlearning algorithm and partially aligns with Boundary Unlearning\cite{Chen_2023_CVPR} or DUCK\cite{cotogni2023duck} that manipulates respectively forget samples labels or forget samples feature vectors to induce unlearning. However, while these methods adjust samples towards incorrect directions based on the nearest wrong class or centroid in feature space, our solution focuses on using the distributions of features to efficiently reorganize the positions of forget samples in the feature space. This enables us to utilize a more precise and efficient mechanism, grounded in the Mahalanobis distance. Nonetheless, SCAR does not rely on any images from the retain set to restore performance during the unlearning process whereas model-agnostic methods like DUCK \cite{cotogni2023duck}, Boundary Unlearning\cite{Chen_2023_CVPR}, and SCRUB\cite{kurmanji2023towards} partially or fully depend on it. SCAR, thanks to an efficient knowledge preservation method based on distillation and OOD data, can operate without accessing training data.

\section{Methods}
\label{sec:methods}

\subsection{Preliminaries}
\label{sec:preliminaries}

Consider a dataset $\{x_i, y_i\}^{N}_{i=1}$ consisting of N images $x_i$ each associated with a label $y_i \in \{0, \dots, K\}$ where $K$ is the number of classes in the dataset. A model $\Phi_{\theta}$ is trained on the training subset of the dataset $\mathcal{D}$ to predict the corresponding label for a given image, that is, $\hat{y_i} = \Phi_{\theta}(x_i)$. This model is then evaluated on the test subset $\mathcal{D}^t$. The objective of a machine unlearning algorithm is to derive a model, denoted as $\Phi_{\theta}^{U}$, from which information about a specific subset of data, known as the forget set $\mathcal{D}_f$, is removed from the original model's parameters $\theta$. Concurrently, the algorithm aims to preserve the model's performance on the retain set $\mathcal{D}_r$, such that it is comparable to that of a model trained exclusively on the retain set. In this paper, we consider two scenarios: Class-Removal (CR) and Homogeneous Removal (HR).

In the CR scenario, the objective is to remove from the model's weights the information about the instances of a class $C$ among the $K$ available. The two subsets $\mathcal{D}$ and $\mathcal{D}^t$ are split in forget ($\mathcal{D}_f$, $\mathcal{D}^t_f$) and retain ($\mathcal{D}_r$, $\mathcal{D}^t_r$) sets. The forget set contains all the images, from the class to be removed  $\mathcal{D}_f=\{x^f_i, y^f_i\}^{N^{f}}_{i=1}$ with $y^f_i \in \{C\}$ while the retain set contains all the remaining images $\mathcal{D}_r=\{x^r_i, y^r_i\}_{i=1}^{N^{r}}$, with $y^r_i \in \{0, \dots, K-1\}$; the same definition stands also for $\mathcal{D}^t$. The desiderata in this scenario is to maintain the original accuracy on the $\mathcal{D}^t_r$ while minimizing the one on the $\mathcal{D}^t_f$. In the HR scenario, the training dataset $\mathcal{D}$ is divided into retain $\mathcal{D}_r$ and forget $\mathcal{D}_f$ sets, while the test set $\mathcal{D}^t$ remains unchanged. In this case, the images constituting $\mathcal{D}_f$ are chosen from all the classes. Here, the goal is to eliminate the model's knowledge of specific training instances in $\mathcal{D}_f$ without compromising its performance on $\mathcal{D}_r$. Following the application of the unlearning algorithm, the model should no longer differentiate between data from $\mathcal{D}_f$ and $\mathcal{D}^t$ or in other words it interprets $\mathcal{D}_f$ samples as never-seen-data. This indistinguishability has also to be reflected in the $\mathcal{D}_f$ accuracy which has to be approximately similar to $\mathcal{D}^t$ accuracy. This second scenario is more challenging compared to CR since $\mathcal{D}_r$ and $\mathcal{D}_f$ include images from the same classes and the model should maintain its ability to generalize across all classes.

\subsection{SCAR}
In SCAR, the unlearning process is based on metric learning and involves directly modifying the feature vector projection of the sample to be forgotten within the feature space. Additionally, we have devised a regularization technique that enables the DNN to preserve the original model information through distillation without access to the retain dataset. By leveraging an OOD dataset, SCAR enables the transfer of knowledge from the original model to the unlearning model, thereby ensuring that classification performance on test samples remains consistent throughout the unlearning process. Initially, we present the unlearning strategy of SCAR, followed by a deeper exploration of the distillation with OOD data and the Metric Learning mechanism. Then we introduce a variant of SCAR, called SCAR self-forget, that works without accessing the forget set.\vspace{1mm}

\minisection{Unlearning strategy}. The unlearning strategy is composed of 2 synergistic mechanisms, \textit{Metric Learning} and \textit{Distillation-Trick} (Fig. \ref{fig:fig1} A). In the following sections each DNN $\Phi_{\theta}$ will be represented as a sequence of a backbone $\Phi_{\psi}$ and a final fully-connected layer $\Phi_{\pi}$ where $\theta,\psi,\pi$ are the corresponding weight. 

During the training phase of the original model $\Phi_{\theta}$, for each class $i^{th}$ the distribution of feature vectors $Q_i = \{ \Phi_\psi(x_j)\}_{j}$ is prototyped\cite{rebuffi2017_proto,xu2020_proto,Lange2021_proto} by its mean $\mu_i$ and covariance matrix $\hat S_Q$ (where $\Phi_\psi(x_j)$ represent a feature vector), and accurately stored preventing any data leakage.
Hence, given a forget sample $(x_j,y_j=k)$ belonging to the $k^{th}$ class, SCAR selects the closest distribution $Q^*_j$ among the set of distribution $Q_i$ with $i \in \{0,...,K\}$ and $ i\neq k$ to the feature vector $\Phi^U_{\psi}(x_j)$.
\begin{equation}
    Q^*_j = \argmin_{Q_i} d_M(\Phi^U_{\psi}(x_j),Q_i)
\end{equation}
Hence, during the unlearning procedure, the algorithm, through the \textit{Metric Learning mechanism}, minimizes the Mahalanobis distance between the forget samples feature vector $\Phi^U_{\psi}(x_j)$ and their corresponding selected distribution $Q^*_j$. The overall forget loss is written in eq. \ref{eq:loss_fgt}:
\begin{equation}
     \mathcal{L}_{M} = \frac{1}{N_{f,\text{batch}}} \sum_{j=0}^{N_{f,\text{batch}}-1} d_M (\Phi^U_{\psi}(x_j),Q^*_j).
     \label{eq:loss_fgt}
\end{equation}
where $N_{f,\text{batch}}$ is the batch size of forget samples.

Since in our proposed scenario, the $\mathcal{D}_r$ dataset is no longer accessible, the algorithm makes use of the \textit{Distillation-Trick mechanism} to retain past knowledge. We used the Jensen-Shannon divergence between the original model $\Phi_\theta$ and the unlearning model $\Phi_\theta^U$ ( see par. \textit{Distillation-Trick}, eq. \ref{eq:trick-distill})
\begin{equation}
    \mathcal{L}_{TD} = \frac{1}{N_{r,\text{batch}}}\sum_{j=0}^{N_{r,\text{batch}}-1}d_{JS}(\Phi^U_{\theta}(x_j)\parallel \Phi_{\theta}(x_j))
    \label{eq:loss_ret}
\end{equation}
where $N_{r,\text{batch}}$ is the batch size of the surrogate retain samples. The overall loss combines the forget and retain contribution in eq. \ref{eq:loss_fgt} and \ref{eq:loss_ret}.
\begin{equation}
    \mathcal{L} = \lambda_1 \mathcal{L}_{M} + \lambda_2 \mathcal{L}_{TD}
    \label{eq:loss}
\end{equation}
where $\lambda_1$ and $\lambda_2$ are hyperparameters.
Instead of selecting a priori the number of unlearning epochs, SCAR  computes the $\mathcal{A}_f$ at each epoch and it stops when $\mathcal{A}_r$ is lower or equal to a threshold $\epsilon$. In CR $\epsilon=0$ because the objective is to remove completely the knowledge of the class to remove. In HR $\epsilon=\mathcal{A}^t$ because the forget test accuracy has to be close to the accuracy of the original model on the test set. When this condition is not reached the unlearning process stops when it reaches the maximum number of epochs allowed ($\text{epochs}_{max}$).\vspace{1mm}

\begin{figure}[t]
    \centering
    \includegraphics{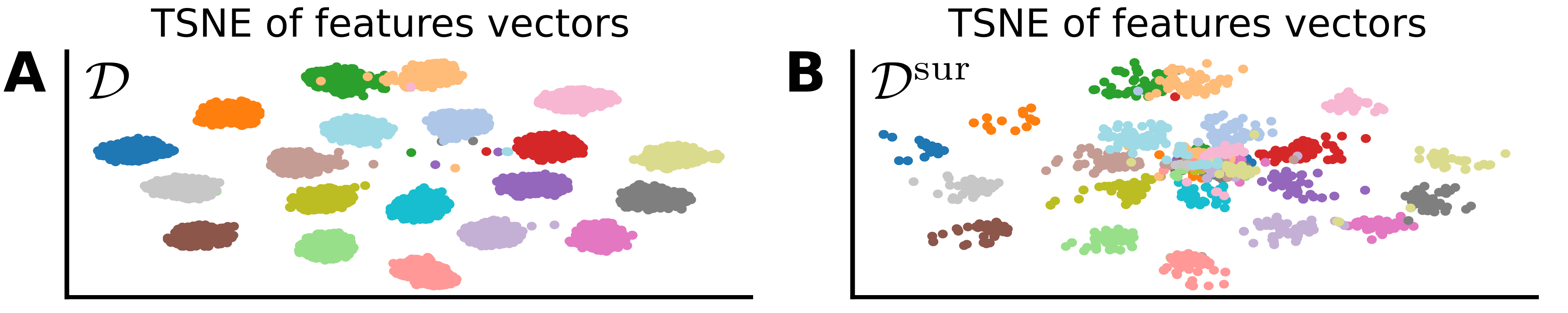}
    \caption{A-B) TSNE plots of feature vectors for the first 20 classes in CIFAR100 (A, $\mathcal{D}$) and the surrogate dataset subset Imagenet (B, $\mathcal{D^{\text{sur}}}$) }
    \label{fig:fig2}
\end{figure}
\minisection{Distillation-trick}. Numerous instances have been documented in literature wherein DNNs exhibit high confidence predictions \cite{guo2017calibration}, even when presented with data significantly distant from their training set. Examples include their performance on fooling images \cite{nguyen2015deep}, OOD samples \cite{hendrycks2017baseline}, or within medical diagnostic tasks \cite{leibig2017leveraging}. For instance, the t-SNE projection of feature vectors from a DNN, both from the original training dataset and an external OOD dataset, exhibited clustering at identical positions within the feature space (Fig. \ref{fig:fig2} A-B). In our proposed algorithm, we leverage this characteristic behavior of DNNs on OOD datasets as a form of regularization during the unlearning process, which we term ``distillation-trick''. This approach involves distilling the knowledge from the original frozen model (referred to as the teacher model) into the unlearning model using an external dataset acting as a surrogate dataset, denoted as $\mathcal{D}^\text{sur}$  \cite{gongfag2021_KDOOD} (Fig. \ref{fig:fig1} B). The distillation-trick leverages the Jensen-Shannon divergence between $\Phi^U_{\theta}(x_j)$ and $\Phi_{\theta}(x_j)$ to retain the knowledge of the original model, even as the unlearning model undergoes training on $\mathcal{D}^\text{sur}$ (omitting the dependency on $x_j$ for brevity):  
\begin{equation}
    d_{JS}(\Phi^U_{\theta}\parallel \Phi_{\theta}/T) = \frac{1}{2}D_{KL}(\Phi^U_{\theta}\parallel \Phi_{\theta}/T) +
     \frac{1}{2}D_{KL}(\Phi_{\theta}/T\parallel \Phi^U_{\theta})
    \label{eq:trick-distill}
\end{equation}
where T is a scaling temperature. To validate this, we trained $\Phi_{\theta}$ using distillation-trick on a surrogate dataset. Our experiments demonstrate that the accuracy of $\Phi_{\theta}$ on $\mathcal{D}_t$ remains consistent across epochs and comparable to that of $\Phi^U_{\theta}$ (refer to Sec. A of the Supp. Materials). \vspace{1mm}

\minisection{Metric learning}. In the context of unlearning problems, metric learning serves as a pivotal concept aimed at modifying the feature vectors of forget samples, thereby altering their class predictions. The prevalent strategies involve shifting the feature vectors of forget samples towards the nearest incorrect class centroids\cite{cotogni2023duck}  or directly shifting the logits towards the nearest incorrect class\cite{Chen_2023_CVPR}. In our algorithm, we propose the application of metric learning by adjusting the feature vectors of forget samples toward the nearest distribution of feature vectors of a different class (Fig. \ref{fig:fig1} C). Thanks to the Mahalanobis distance, which gauges the distance between the forget set samples and the whole distributions of feature vectors, SCAR can integrate more comprehensive information about the distributions through the covariance matrix. This approach contrasts with simpler methods that represent distributions solely using centroids (see Sec.\ref{sec:measure_comparison} for a comparison with other distances based on centroids).
Therefore, given the set of distribution of feature vectors for each class $Q = \{ Q_i\}_{i=0}^{K}$ the Mahalanobis distance is defined in eq. \ref{eq:Mahalanobis}.
\begin{equation}
    d_M(\Phi^U_\psi(x_j),Q_i) = \sqrt{(\Phi^U_\psi(x_j)-\mu_Q)^{T}\hat{S}_Q^{-1}(\Phi^U_\psi(x_j)-\mu_Q)}
    \label{eq:Mahalanobis}
\end{equation}
where $\Phi^U_\psi(x_j)$ is the feature vector of $x_j$ (i.e. the output of the backbone of $\Phi^U$), $\mu_Q$ and $\hat{S}_Q$ are the mean and the normalized covariance matrix obtained from the distribution $Q_i$.

Each class's covariance matrix $S_Q$ will exhibit varying levels of scaling and variances across different dimensions. Consequently, the Mahalanobis distances of features from distinct classes will be subjected to differing scaling factors. Therefore, to standardize the covariance matrices per class effectively, we conduct correlation matrix normalization on all the covariance matrices (eq. \ref{eq:normalization}):
\begin{equation}
    \hat{S}_Q(i,j) = \frac{S_Q(i,j)}{\sigma_Q(i)\sigma_Q(j)}, \quad \sigma_Q(j)=\sqrt{S_Q(j,j)}, \quad \sigma_Q(i)=\sqrt{S_Q(i,i)}.
    \label{eq:normalization}
\end{equation}
Since the number of samples per class in public datasets ($\approx$ 500 for CIFAR100 and TinyImagenet) is generally lower than the number of dimensions in the feature space (e.g. N=512 for \texttt{resnet18}), we use a covariance shrinkage method to get a full-rank matrix eq. \ref{eq:mat_shrink}\cite{Chen2009Shrinkage}. 
\begin{equation}
    S_Q^{s} = S_Q +\gamma_0 V_0 \mathbb{I} +\gamma_1 V_1 (1-\mathbb{I})
    \label{eq:mat_shrink}
\end{equation}
where $V_0$ is the average diagonal variance, $V_2$ is the average off-diagonal covariance of $S_Q$ and $\mathbb{I}$ is the identity matrix.
Furthermore, we added Tukey’s normalization \cite{ttukey77} to the feature vectors $\Phi^U_\psi(x)^{\delta}$ to obtain approximately Gaussian features, where $\delta$ is a hyperparameter to decide the degree of transformation of the distribution.\vspace{1mm}

\vspace{1mm}

\subsection{SCAR Self-forget}
\begin{wrapfigure}{l}{6.4cm}
\vspace{-.2cm}
\centering
\includegraphics[width=0.5\textwidth]{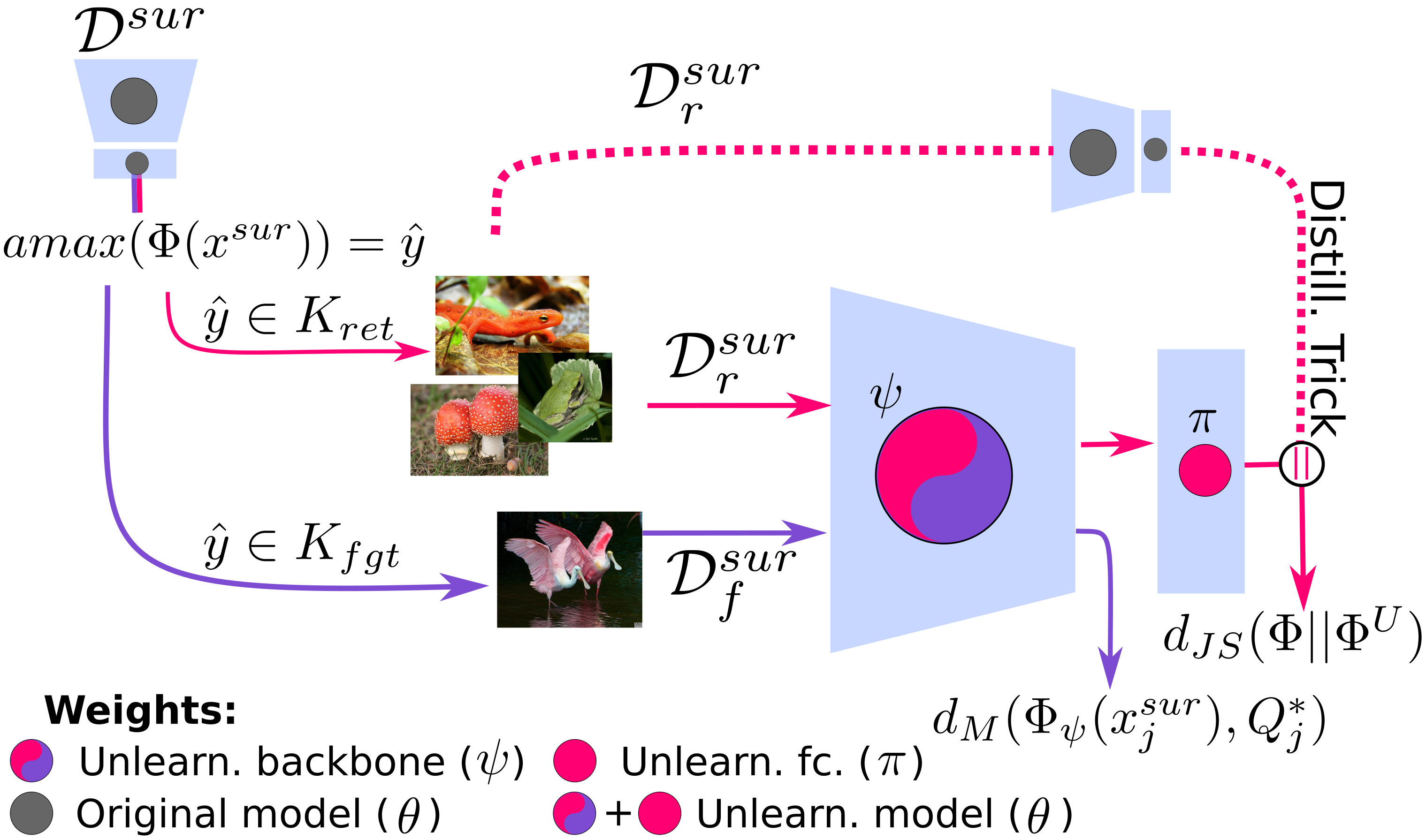}
\vspace{-0.0cm}
\caption{Scheme of SCAR self-forget in CR. The surrogate dataset $\mathcal{D}^\text{sur}$ supplies during the unlearning procedure both the surrogates $\mathcal{D}_r^\text{sur}$ and $\mathcal{D}_f^\text{sur}$.}
\vspace{-0.4cm}
\label{fig:self-forget}
\end{wrapfigure}

Additionally, we present an adaptation of SCAR tailored for a more demanding CR scenario, wherein access to the forget set is restricted, and only the class ID slated for removal is provided. We term this procedure SCAR self-forget (Fig. \ref{fig:self-forget}). In this version, both the forgetting mechanism and the distillation-trick rely on the surrogate dataset. Given $K_\text{ret}$, the set of retained classes, and $K_\text{fgt}$, the set of forget classes, SCAR initially employs the original model $\Phi_\theta$ to classify $\mathcal{D}^\text{sur}$.
Subsequently, SCAR splits the samples in $\mathcal{D}^\text{sur}$ into $\mathcal{D}^\text{sur}_r$ and $\mathcal{D}^\text{sur}_f$ if respectively the predicted class $\hat{y}$ belongs to $K_\text{ret}$ , and in $\mathcal{D}^\text{sur}_f$ if $\hat{y}$ is within $K_\text{fgt}$. 
Thus, $\mathcal{D}^\text{sur}_f$ serves as a surrogate for $\mathcal{D}_f$, and SCAR is optimized using the loss function in Eq. \ref{eq:loss}. This configuration poses an extreme challenge as it lacks both retain and forget data during the unlearning process. SCAR self-forget marks the first attempt to address the unlearning problem within the Class-Removal (CR) scenario without necessitating training or forget data, nor the retention of specific information during training such as gradient updates.

\section{Experimental Results}
\label{sec:exps}

We evaluated SCAR against a range of sota methods and baselines on three datasets CIFAR10\cite{krizhevsky2009learning}, CIFAR100\cite{krizhevsky2009learning}, and TinyImageNet\cite{le2015tiny}. CIFAR-10 and CIFAR-100 are comprised of 60,000 images of 32$\times$32 pixels each (50,000 images for train and 10000 for test), with CIFAR-10 divided into 10 classes, and CIFAR-100 categorized into 100 classes. TinyImageNet consists of 110,000 images of 64$\times$64 pixels each, divided into 200 classes (100,000 for train and 10,000 for test). The considered sota methods include DUCK\cite{cotogni2023duck}, Boundary Shrink\cite{Chen_2023_CVPR}, Boundary Expanding\cite{Chen_2023_CVPR}, SCRUB\cite{kurmanji2023towards}, L1-Sparse\cite{jia2023model}, Bad Teacher\cite{chundawat2023can}, ERM-KTP\cite{lin2023erm}, Negative Gradient\cite{golatkar2020eternal}, and Random Labels\cite{hayase2020selective}, alongside traditional approaches such as Retraining and Fine-Tuning. While the latter two are computationally demanding as they necessitate access to the $\mathcal{D}_r$, they serve as critical benchmarks. Retraining is by definition the upper bound for unlearning, removing the $\mathcal{D}_f$ from the training process entirely. Fine-Tuning, conversely, adjusts the pre-trained model weights using only the $\mathcal{D}_r$.
Importantly, changing the number of epochs can substantially affect the finetuning performance: few epochs result in an efficient unlearning whereas if the model is finetuned for many epochs it is substantially retrained. For this reason, the original model is finetuned for a sufficient number of epochs ($\ll$ epochs of retrain) that guarantees to unlearn $\mathcal{D}_f$ (as outlined in \cite{Chen_2023_CVPR,lin2023erm}). Unfortunately, for this reason, Fine-Tuning is extremely time-consuming and inefficient compared to other sota methods \cite{Chen_2023_CVPR,lin2023erm,cotogni2023duck}. All methods utilized \texttt{resnet18}\cite{he2016deep} as the neural architecture.
The details about the hyperparameters are reported in the Supp. Material, Sec. \textcolor{Red}{B}, "Reproducibility".

\begingroup
\setlength{\tabcolsep}{7pt} 
\renewcommand{\arraystretch}{1.1} 
\begin{table}[t]
    \caption{Comparison between SCAR performance and different surrogate dataset in CR scenario. The metrics are reported as mean $\pm$ std over ten runs.}
    \centering
    \label{tab:surrogate}
    \input{tables/table1}
\end{table}
\endgroup

We established two experimental frameworks, CR and HR, based on the scenarios outlined in Sec. \ref{sec:preliminaries}. For CR, we applied a constant seed, dividing each dataset into $\mathcal{D}_r$ and $\mathcal{D}_f$ with distinct class distributions. The $\mathcal{D}_f$ included instances from a single class, while the $\mathcal{D}_r$ comprised the remaining classes. This setup was replicated across ten splits to calculate average metrics and standard deviations. In the HR scenario, we selected 10 different seeds for dataset partitioning, ensuring a 90:10 split between $\mathcal{D}_r$ and $\mathcal{D}_f$, with class overlaps but unique instances. We reported the mean and standard deviation of the results across these seeds. Further details are available in the Sec. B of the Supp. Material.

For the CR scenario, the metrics considered are the accuracies on retain ($\mathcal{A}^t_r$) and forget ($\mathcal{A}^t_f$) test sets,  to be maximized and minimized, respectively, and the Adaptive Unlearning Score (AUS)\cite{cotogni2023duck}. This final score facilitates the comparison among various methods by encapsulating forget and retain test set performances into a single metric. It captures a delicate balance between two essential objectives: preserving high test accuracy while effectively tackling the unlearning task (see Supp. Material Sec. B).
For the HR scenario, the metrics included are accuracy on $\mathcal{D}_f$ ($\mathcal{A}_f$), overall test set accuracy ($\mathcal{A}^t$), AUS, and the success of a membership inference attack (MIA)\cite{Salem2019,song2020,cotogni2023duck}, with the optimal MIA value being $\mathcal{F}_1=0.5$. Specifically the number of forget and test samples are balanced and $\mathcal{F}_1=0.5$ (chance level) means the impossibility to distinguish the membership of the forget data. The Sec. B of the Supp. Material provides details about hyperparameters for MIA tuning.

\subsection{Impact of different Datasets as $\mathcal{D}^\text{sur}$ }

A critical role in SCAR is played by the distillation-trick mechanism. It allows SCAR to maintain acquired information about the retain set even without accessing it during the unlearning. The distillation-trick works using an OOD dataset, and investigating how different datasets can or cannot affect the efficiency of SCAR is of utmost importance. For this reason, following ~\cite{bibas2021single,fort2021exploring,hsu2020generalized}, we selected 4 different OOD external public datasets characterized by different distributions of pixels, number of images, resolution, and number of classes (a random subset of 10K images of Imagenet1K \cite{deng2009imagenet}, a random subset of 10K images of COCO \cite{lin2014microsoft}, a set of 2K public natural images\cite{randomIMG}, and a set of 10K images distilled with \cite{yin2024squeeze}  from Imagenet1K). We verified through the Kolmogorov-Smirnov (KS) test \cite{massey1951kolmogorov} that the distributions of pixels of these surrogate datasets are different from the datasets used for benchmarking SCAR (Tab. \ref{tab:surrogate} p-val columns). This is fundamental to assess the generality of SCAR and it does not rely on data leakage between datasets; this is especially true for tinyImagenet and the random subset of Imagenet1K where we observed the KS p-val is <0.001 . Moreover, we also used a surrogate dataset made of Gaussian noise images where the semantic information was absent. We performed our experiments in the CR scenario using CIFAR100 as $\mathcal{D}$. Both accuracies and AUS scores were statistically comparable across datasets (Tab. \ref{tab:surrogate}) even if the Imagenet1K subset gives the overall highest results. Importantly, when in the surrogate dataset semantic information is removed (Gaussian Noise) SCAR performance is poor, suggesting how the surrogate dataset has to contain some spatial feature and patterns to trigger the DNN filters. Hence, the presence of semantic information content in images plays a pivotal role in the distillation-trick mechanism, but, at the same time, the differences in terms of kind of information (e.g. classes and distribution of pixels) do not affect this technique.
For the rest of the paper results are presented using as a surrogate the OOD random subset of Imagenet1K.

\begingroup
\setlength{\tabcolsep}{7pt} 
\renewcommand{\arraystretch}{1.1} 
\begin{table}[t]
    \caption{Performance comparison between SCAR and sota methods in CR scenario for CIFAR10, CIFAR100, and TinyImagenet datasets. The metrics are reported as mean $\pm$ std over ten runs. Results marked with $^\dagger$ are taken from \cite{cotogni2023duck}. Columns ``$\mathcal{D}_r$ free'' and ``$\mathcal{D}_f$ free'' indicate whether the method was (\cmark) or not (\xmark), Retain-Free or Forget-free, respectively.  \textcolor{green}{Green} and \textcolor{red}{red} colors highlight respectively the best results between methods based on retain set and between retain free methods.}
    \vspace{-2mm}
    \input{tables/table2}
    \label{tab:CR}
\end{table}
\endgroup
\begingroup
\setlength{\tabcolsep}{4pt} 
\renewcommand{\arraystretch}{1.1} 
\begin{table}[t]    
    \caption{Performance comparison between SCAR and sota methods in HR scenario for CIFAR10, CIFAR100, and TinyImagenet datasets. The metrics are reported as mean $\pm$ std over ten runs. Results marked with $^\dagger$ are taken from \cite{cotogni2023duck}. Columns ``$\mathcal{D}_r$ free'' indicate whether the method was (\cmark) or not (\xmark), Retain-Free. \textcolor{green}{Green} and \textcolor{red}{red} colors highlight respectively the best results between methods based on retain set and between retain free methods.}
    \centering
    \vspace{-2mm}
    \input{tables/table3}
    \label{tab:HR}
\end{table}
\endgroup
\subsection{Comparison with sota methods}

We evaluated SCAR, and SCAR Self-Forget, within the CR scenario across three datasets (Tab. \ref{tab:CR}). Our findings indicate that our method surpasses Retain-Free approaches like Negative Gradient and Random Label in performance. Furthermore, \textbf{SCAR's performance is on par with methods that leverage both forget and retain sets}. Notably, SCAR's performance is lower than Fine-Tuning and within 1 std. from SCRUB in CIFAR100 and TinyImagenet, though these methods depend heavily on the $\mathcal{D}_r$ requiring several finetuning steps on it. Importantly, these outcomes are also achieved with SCAR Self-Forget, which remarkably does not utilize any data from the original training set. 

We also extended our analysis to the HR scenario (Tab. \ref{tab:HR}). Importantly, the Self-Forget variant of our method is not applicable in HR, as the $\mathcal{D}_f$ is essential to identify the specific instance to forget. In this more demanding context, \textbf{SCAR surpasses all the performance of Retain-Free approaches}. Between the methods that use the $\mathcal{D}_r$ only DUCK in CIFAR100 (within 1 std.) and TinyImagenet and Fine-Tuning in TinyImagenet achieve superior results. 

Notably, our method outperforms the Retain-Free approaches in both scenarios and demonstrates commendable performance relative to methods that rely on the $\mathcal{D}_r$. This distinction highlights the effectiveness and versatility of our approach, making it a viable alternative for scenarios where the $\mathcal{D}_r$ is entirely unavailable. We also confirmed our result using the COCO dataset as a surrogate dataset (Sec. A of the Supp. Material).

\begingroup
\setlength{\tabcolsep}{7pt} 
\renewcommand{\arraystretch}{1.1} 
\begin{table}[t]
    \centering
    \caption{Results of the ablation study on CIFAR100 in CR and HR scenarios. The metrics for CR and HR are reported as mean $\pm$ std over ten runs.}
    \vspace{-2mm}
    \input{tables/table4}
    \label{tab:ablation}
\end{table}
\endgroup

\subsection{Ablation Study}

We conducted an ablation study to evaluate the impact of different loss components $\mathcal{L}_{LD}$ and $\mathcal{L}_{M}$ of our method on the unlearning performance across CR and HR scenarios. The results of this study are presented in Tab. \ref{tab:ablation}. 
In CR, considering only $\mathcal{L}_{LD}$ leads to similar retain set accuracy but with high $\mathcal{D}_f$ accuracy. In fact, the stopping criteria is hardly fulfilled and the method stops at the maximum number of epochs allowed. Conversely, considering only $\mathcal{L}_{M}$ the methods can push down the accuracy on the $\mathcal{D}_f$, but at the cost of a low retain set accuracy. In the HR scenario, we observed similar but more pronounced results. The presence of $\mathcal{L}_{LD}$ alone is insufficient for effectively erasing knowledge about forget samples, as evidenced by a high forget accuracy. Employing only $\mathcal{L}_{LD}$ results in a steep decrease in both accuracies. This result suggests how the unlearning strategy without a method for retaining knowledge can't be used to unlearn. This effect is more evident in HR because images from the same class are distributed across both the $\mathcal{D}_r$ and $\mathcal{D}_f$, meaning that the $\mathcal{L}_{LD}$ compromises the general knowledge of the classes. 

\begingroup
\setlength{\tabcolsep}{6pt} 
\renewcommand{\arraystretch}{1.1} 
\begin{table}[t]
    \centering
    \caption{Results of SCAR changing the measure used in $\mathcal{L}_{M}$  on the CIFAR100. The metrics for CR and HR are reported as mean $\pm$ std over ten runs.}
    \vspace{-2mm}
    \input{tables/table5}
    \label{tab:distances}
\end{table}
\endgroup

\begin{minipage}[!th]{0.95\textwidth}
  \begin{minipage}[b]{0.49\textwidth}
    \centering
    \includegraphics[]{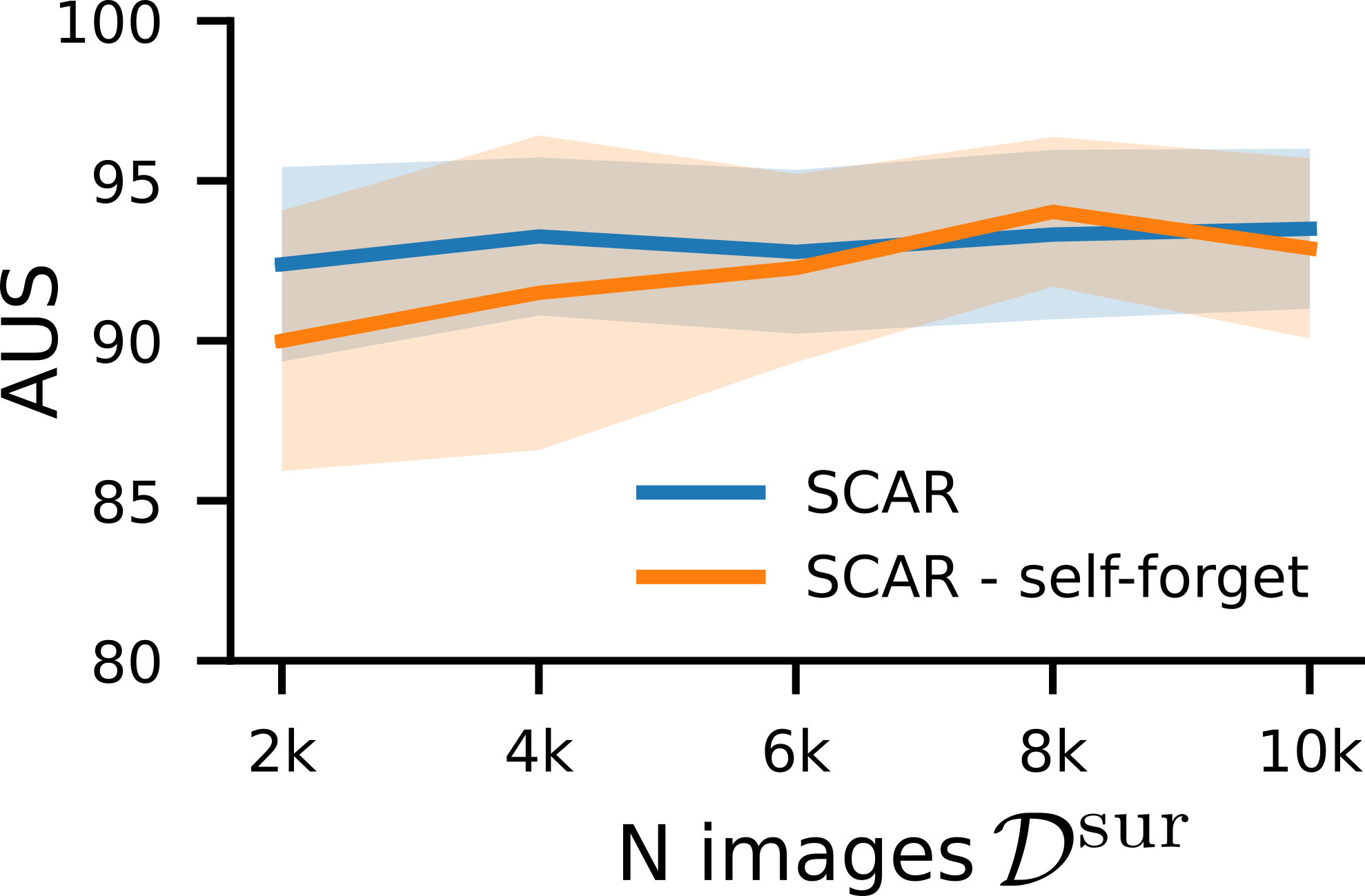}
    \label{fig:size}
    \captionof{figure}{Values of final AUS for SCAR and SCAR self-forget as a function of the number of samples available in $\mathcal{D}^{\text{sur}}$ (subset of Imagenet1K). AUS is reported as mean $\pm$ std over ten runs }
  \end{minipage}
  \hfill
  \begin{minipage}[b]{0.49\textwidth}
    \centering
    \begingroup
    \setlength{\tabcolsep}{6pt} 
    \renewcommand{\arraystretch}{1.1} 
    \input{tables/table6}
    \label{tab:architecture}
      \captionof{table}{Results of SCAR on CIFAR100 dataset in the CR scenario changing the model to unlearn. The metrics are reported as mean $\pm$ std over ten runs before and after the unlearning.}
    \endgroup
    \end{minipage}
  \end{minipage}

\subsection{Analysis on the effect of different measures in the metric learning mechanism of SCAR}
\label{sec:measure_comparison}

We report in Tab. \ref{tab:distances} the results of the SCAR using different distance measures for the $\mathcal{L}_{M}$ component across CIFAR100 and TinyImagenet datasets in both the CR and HR scenarios. Using the Mahalanobis distance yields a perfect balance, achieving the highest AUS scores in both CR and HR. Cosine Similarity distance achieves compatible results (even if constantly lower) in CR and HR scenarios but at the cost of being ($22\pm5 \%$) slower than the Mahalanobis. This result suggests how the information about the distributions of samples embedded into the Mahalanobis distance is fundamental to guide the optimization efficiently. In contrast, the $L_2$ distance resulted in lower performance in both CR and HR scenarios suggesting how this metric hampers the unlearning process.

\subsection{Analysis of the impact of the dimension of $\mathcal{D}^\text{sur}$}

Fig.~\textcolor{red}{4} illustrates the relationship between the number of images in the surrogate dataset $D_{sur}$ (subset of ImageNet1K) and the AUS for both SCAR and its self-forget variant, in the CR scenario. SCAR maintains a relatively stable AUS score as the number of images increases, suggesting that the method's unlearning ability does not significantly depend on the size of the surrogate dataset. The self-forget variant of SCAR shows a plateau in AUS score when at least 6K surrogate images are used which suggests how the dimension of $\mathcal{D^{\text{sur}}}$ can affect the self-forget mechanism. Importantly, the self-forget mechanism that can access 6K surrogate samples or more exhibits performance comparable with the method that can access forget data. This pattern reflects the effectiveness of SCAR in handling the unlearning process, with only minor improvements seen with the addition of more surrogate data when forget data are not available.

\subsection{Architectural-Agnostic Results}
For additional insight, we examined the performance of SCAR (Tab. ~\textcolor{red}{6}) across various DNN architectures within the CR scenario. We investigated the effect of the unlearning with SCAR on a basic CNN network (AllCNN), ResNet architectures (18, 34, and 50), and a Vision Transformer (ViT-B16). Our method consistently demonstrates strong performance across all tested architectures. Notably, the results achieved with the ViT confirm the model-agnostic nature of our approach beyond the CNN architectures. It is worth mentioning that the efficacy of our method is not contingent on the initial accuracy of the model, which underscores its adaptability and broad applicability.

\section{Conclusions}
\label{sec:concl}
In this paper, we introduced SCAR a novel method for approximate unlearning, designed to effectively erase information about the forget-set data while maintaining the original model performance. Our experimental results indicate that SCAR outperforms the methods that do not require access to the retain set and achieves performance levels comparable to, or even surpassing, those of existing approximate unlearning methods that depend on the retain set, across both CR and HR scenarios. Furthermore, we have developed a self-forget variation of SCAR capable of delivering impressive results in CR without accessing both the retain and forget sets. A possible limitation of our approach is a marginal decrease in accuracy compared to the original model, a consequence of the inability to access either the forget or retain sets. Like many approximate unlearning methods, future work should include a mathematical exploration of the algorithms' certifiability. 

Despite these challenges, we firmly believe that our algorithm, that do not require access to any retain set, can significantly impact society: for example in the performance restoration of models poisoned by corrupted data or in recognition systems (such as facial recognition), by enabling individuals to exercise their right to have their data removed from the model's knowledge.
\appendix
\section*{Supplementary Material}
\section{Additional Analyses} 
In this section, we present additional analyses that support the findings of the main paper. Specifically, we explore the impact of the distillation trick beyond the unlearning context. Additionally, we display the results for both CR and HR scenarios using the COCO dataset as $\mathcal{D}^{sur}$. We also examine the computational impact of our method. Lastly, we investigate the behavior of feature maps from various layers of the original and unlearned models in the feature space.
\begin{figure}
    \centering
    \includegraphics{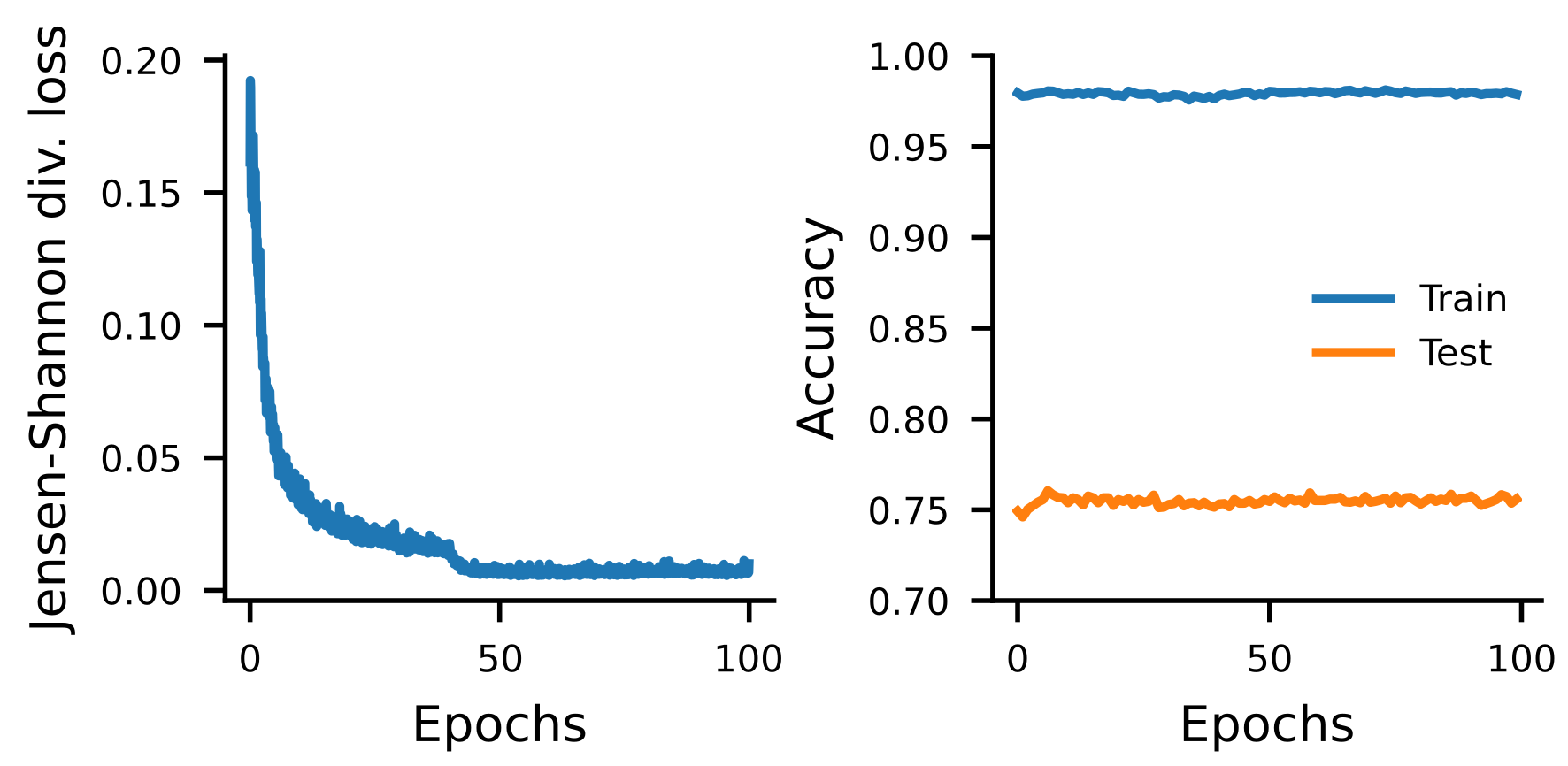}
    \caption{Examples of train loss (left) and train set and test set accuracies (right) of a \texttt{resnet18} trained with the trick-distillation mechanism on the subset of Imagenet1K.}
    \label{fig:trick-distil}
\end{figure}
\subsection{Distillation-Trick details}
The distillation-trick mechanism has been verified outside the unlearning scope. We used as $\phi_\theta$ a $\texttt{resnet18}$ trained on CIFAR100 and as OOD surrogate dataset the subset of 10K images from Imagenet1K. $\phi_\theta$ has been trained using the distillation-trick mechanism (eq. 1 and 7) directly on the surrogate dataset and tested on the test set of CIFAR100 (Fig. \ref{fig:trick-distil}). For this training, it has been used  Adam\cite{loshchilov2017decoupled} as the optimizer, a weight decay of $5\times10^{-4}$, $lr= 5\times 10^{-4}$ and StepLR scheduler with $\gamma=0.1$ and $\text{epoch}_{\text{step}}=40$.
Overall, we observed that the accuracy of the test set after the first 5 epochs remains constant at $\approx76\%$. This result demonstrates how this mechanism can be used as Model ``Knowledge'' regularization technique when training data are not available as in the CR and HR scenarios where the $\mathcal{D}_r$ is not available.

\subsection{Results with Coco as $\mathcal{D}^{\text{sur}}$}
We reported in Tab. \ref{tab:COCO_sur} the results obtained for SCAR and SCAR self-forget in CR and SCAR in HR for both CIFAR100 and TinyImagenet using the subset of COCO as surrogate dataset. Overall, we observed compatible results in both scenarios which highlights once more how differences in terms of kind of information  such as surrogate dataset classes and distribution of pixels do not affect SCAR.  
\begingroup
\setlength{\tabcolsep}{7pt} 
\renewcommand{\arraystretch}{1.1} 
\begin{table}[]
    \centering
    \input{tables/table_1_suppl}
    \caption{Performance of SCAR and Scar self-forget in CR and HR scenarios for CIFAR100 and TinyImagenet datasets using either the subset of Imagenet1K or the subset of COCO as surrogate datasets. The metrics are reported as mean $\pm$ std over ten runs. S-F stands for Self-forget}
    \label{tab:COCO_sur}
\end{table}
\endgroup

\subsection{Computational Analysis}
We present in Table \ref{tab:time} a computational comparison between our method and other state-of-the-art approaches. We evaluated the methods based on the AUS metric, required unlearning time, and memory usage for the forget and retain data. Notably, our method only needs to store $\mathcal{D}^\text{sur}$ for the Self-Forget variant while $\mathcal{D}_f$ and $\mathcal{D}^\text{sur}$ for SCAR. Furthermore, our approach demonstrates a favorable balance between AUS, unlearning time, and storage efficiency, outperforming several methods that rely on both $\mathcal{D}_r$ and $\mathcal{D}_f$. Although slower than Retain-Free approaches and requiring marginally more memory, our methods exhibit significantly superior performance compared to these approaches and are comparable to more computationally intensive methods that depend on the Retain set.
\begin{table}[]
    \centering
    \input{tables/table_2_supp}
    \caption{Comparison between our method, the self-forget variant, and methods from the state-of-the-art on the Cifar100 dataset in the CR scenario in terms of AUS, unlearning time, and memory requirements. For SCAR and SCAR Self-Forget, the amount of memory occupied for the retain-data refers to the $\mathcal{D}^{\text{sur}}$}
    \label{tab:time}
\end{table}

\subsection{t-SNE analysis on DNN layers}

\begin{figure}
    \centering
    \includegraphics{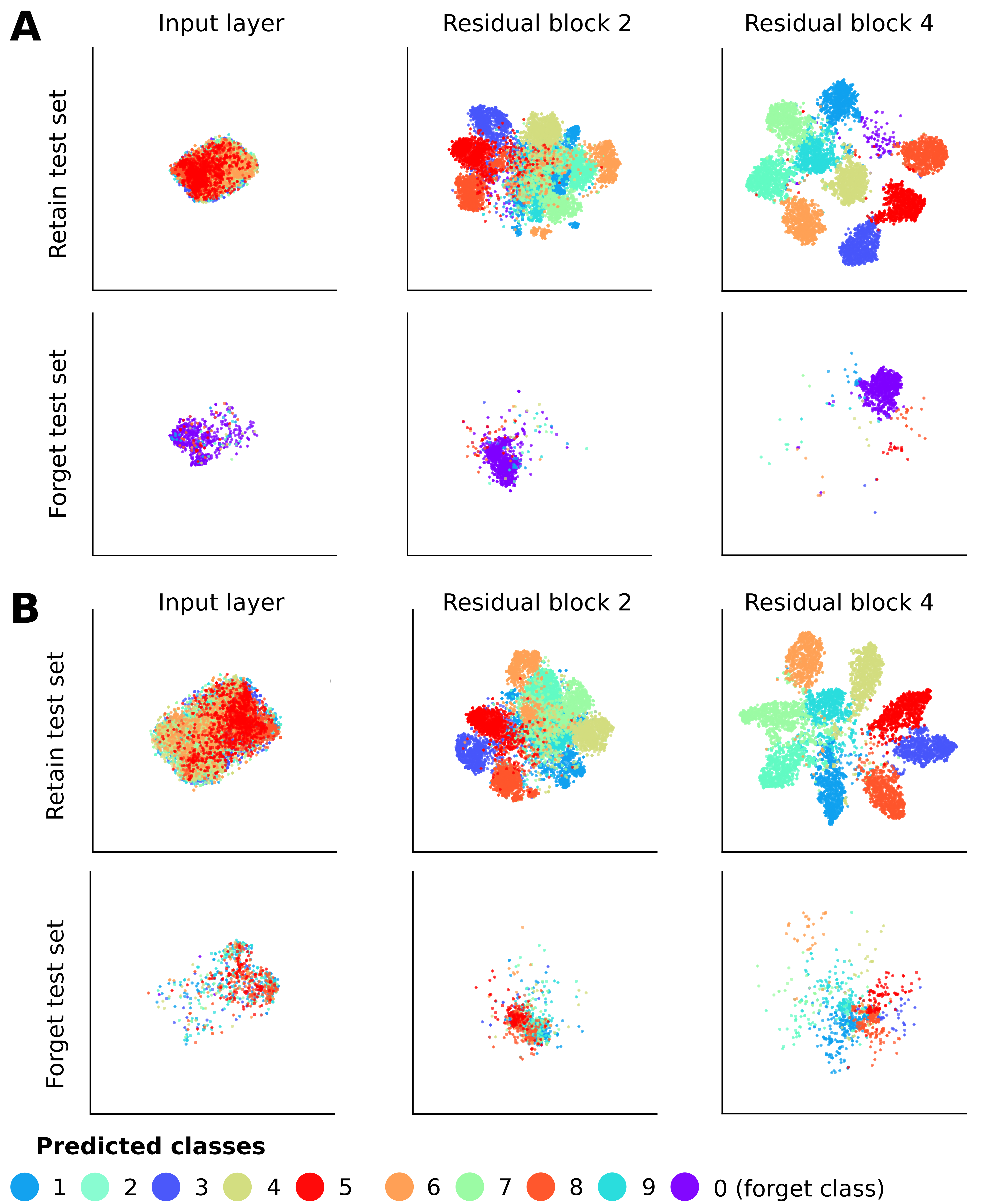}
    \caption{T-SNE plot of feature vectors of the original model $\phi_\theta$ (A) and unlearned model $\phi^U_\theta$ (B) applied to $\mathcal{D}^t_r$ and $\mathcal{D}^t_f$ of CIFAR10. Feature vectors are extracted after the first convolutional layer, and after the first and last residual block. The original model or unlearned model predicted classes are reported in different colors.}
    \label{fig:tsne_residualBlock}
\end{figure}

To gain deeper insights into the impact of the unlearning mechanism employed in our approach, we conducted an analysis of the embeddings derived from the neural network backbone ($\Phi_{\theta}$) both pre and post the unlearning process in the CR scenario. Utilizing t-SNE \cite{van2008visualizing}, a dimensionality reduction technique for high-dimensional data visualization, we analyzed the transformation of embeddings from various layers of the deep neural network (DNN). The resulting distributions, as illustrated in Fig. \ref{fig:tsne_residualBlock}, depict each class within the dataset as a uniquely colored cluster for better visual distinction.

We generated two separate t-SNE visualizations: one representing the original model's embeddings (Fig. \ref{fig:tsne_residualBlock}-A) and the other post-unlearning with SCAR (Fig. \ref{fig:tsne_residualBlock}-B). Initially, in the original model, the embeddings of different classes, particularly from the last residual layer, formed distinct and cohesive clusters, signifying a high degree of separation within the test set. Contrastingly, after applying SCAR, while the embeddings of the samples meant to be retained preserved their clustered integrity, those designated for forgetting became interspersed among the clusters of other classes.

These findings suggest that the information content of the embeddings derived from $\Phi_{\theta}$ for the forget-set no longer hold the necessary attributes for accurate sample classification, signaling a comprehensive dilution of relevant details not only in the final layer but also across preceding layers of the network. This observation underscores the profound efficacy of our unlearning process, which effectively destroys the pertinent data footprint throughout the neural architecture.

\section{Reproducibility}
In this section we report all the details about the experiments reported in the main paper and in this Supplementary Material.

\subsection{Experimental Details and Hyperparameters}

We trained the original model with the seed fixed to $S=42$. Then for the two scenarios proposed in the paper, we adopted two different setups.

\minisection{CR}. In CR scenario, we set the seed to $S=42$. Subsequently, we divided the training and testing datasets into forget sets (comprising all instances of a specific class $q$) and retain sets (comprising instances of the remaining classes). We applied the unlearning algorithms to these sets and recorded the metrics. This process of splitting and applying the unlearning procedure was repeated ten times, each time altering the class $q$ designated for the forget set. Ultimately, we calculated the mean and standard deviation (std) for all metrics. For CIFAR10, one class at time constituted the forget sets ($q=[0,1,2...]$). For CIFAR100, the forget set contained at each time a class multiple of 10, starting from class 0 ($q=[0,10,20,...,90]$). Similarly, for TinyImagenet, starting from class 0, at each iteration, a class multiple of 20 constituted the forget set ($q=[0,20,40,60,...,180]$)

\minisection{HR}. In the HR scenario, we select a seed $S$ and split the training dataset into forget set (containing 10\% of the training data independently from the class) and retain set (containing the remaining 90\%). We applied the unlearning algorithms to these sets and recorded the metrics. This process of splitting and applying the unlearning procedure was repeated ten times, each time selecting a different seed in $S=[0,1,2,3,4,5,6,7,8,42]$ designated for the forget set. Ultimately, we calculated the mean and standard deviation (std) for all metrics.

All the hyperparameters of SCAR used to produce the results reported in Tab.~\textcolor{red}{1} and \textcolor{red}{2} of the main text are detailed in Tab. \ref{tab:hyper_scar}, while the ones used for SCAR self-forget case are listed in Tab. \ref{tab:hyper_selff}. \textbf{For comprehensive reproducibility, all scripts necessary to replicate our findings are attached to this Supplementary Material} and will be released at \url{https://github.com/jbonato1/SCAR}.
\begingroup
\setlength{\tabcolsep}{4pt} 
\renewcommand{\arraystretch}{1.1} 
\begin{table}[]
    \centering
    \input{tables/table_3_supp}
    \caption{Hyperparameters of SCAR used to produce the results reported in Tab. \textcolor{red}{1} and Tab. \textcolor{red}{2} of the main text for the two scenarios, CR on the left, HR on the right.} 
    \label{tab:hyper_scar}
\end{table}
\endgroup

\begingroup
\setlength{\tabcolsep}{5pt} 
\renewcommand{\arraystretch}{1.1} 
\begin{table}[]
    \centering
    \input{tables/table_4_supp}
    \caption{Hyperparameters of SCAR self-forget used to produce the results reported in Tab. \textcolor{red}{1} of the main text, for the CR scenario.}
    \label{tab:hyper_selff}
\end{table}
\endgroup
\subsection{AUS}

The AUS metric \cite{cotogni2023duck} has been formulated to address the need for a measure that considers both the performance on the retain and forget-set of an untrained model. It serves the pivotal function of enabling the ranking of different methods based on their ability to balance memory retention and forgetting. This metric is particularly designed to provide a singular score reflecting each method's efficiency in maintaining high test set accuracy while successfully engaging in the unlearning process. The metric is outlined as follows:
\begin{equation}
    \text{AUS} = \frac{1-(\mathcal{A}^{Or}_{t}-\mathcal{A}_{t})}{1 + \Delta}, \hspace{1em}    \Delta = 
    \begin{cases}
      | 0 - \mathcal{A}_{f}| & \text{if CR}\\
      | \mathcal{A}_{t} - \mathcal{A}_{f}| & \text{if HR} 
    \end{cases},
    \label{eq:AUS}
\end{equation}
In this formulation, $A_{t}$ represents the accuracy on the test set and $A_{f}$ represents the accuracy on the forget-set of the untrained model, whereas $A^{Or}_t$ denotes the accuracy on the test samples of the original model. To simplify notation, we sobstituted $\mathcal{A}_f$ with $\mathcal{A}^t_{f}$, $\mathcal{A}^{Or}_t$ with $\mathcal{A}^{t,Or}_r$, and $\mathcal{A}_t$ with $\mathcal{A}^t_r$. By offering a unified score, the AUS metric simplifies the comparative assessment of different models, underscoring their performance in both retaining essential information and effectuating targeted unlearning.

\subsection{MIA}
In the HR scenario, through the usage of a membership inference attack (MIA)\cite{cotogni2023duck}, the goal is to investigate whether forget set samples were used to train an input model $\Phi_{\theta}$.  To achieve this, the MIA tries to assess whether, from the model perspective, the forget data can be distinguished from the test set data (the actual set of data never seen by the model $\Phi_{\theta}$ during training). Failure of the MIA indicates the inability to differentiate forget data from test data and consequently the impossibility to classify the former samples as training data.\\

Summarizing the step of the MIA proposed in \cite{Salem2019,song2020,cotogni2023duck}: given the datasets $\mathcal{D}_f$ and $\mathcal{D}^t$, we combine $\mathcal{D}_f$ and an equal number of samples from  $\mathcal{D}^t$ into a single dataset, which we then split into a training set $\mathcal{D}_{\text{mia}}$ (containing 80$\%$ of the samples) and a test set $\mathcal{D}_{\text{mia}}^{t}$ (containing 20$\%$ of the samples).\\
Subsequently, an SVM with a Gaussian kernel classifier is trained. This SVM takes as input the probability distribution across classes obtained from the $\texttt{softmax}(\Phi^U_{\theta}(x_i))$, where $x_i$ belongs to either $\mathcal{D}_{\text{mia}}$ or $\mathcal{D}_{\text{mia}}^{t}$. The SVM is trained on $\mathcal{D}_{\text{mia}}$ to distinguish between training (i.e. forget) and test instances.\\
SVM hyperparameters are optimized through a 3-fold cross-validation grid search and finally, the SVM is evaluated on $\mathcal{D}_{\text{mia}}^{t}$. Failure of the Membership Inference Attack (MIA) indicates that information about forget-samples has been successfully removed from the untrained model. The MIA performance is assessed based on the mean F1-score over 10 iterations of the SVM, each utilizing different train and test splits of $\mathcal{D}_f$ and $\mathcal{D}^t$.\\

In the CR scenario, the forget set consists solely of samples from a single class, differently from the HR scenario where the test set is composed of instances from multiple classes. In this scenario, the application of MIA as defined before, would lead the SVM to detect biases in the logits related to the identity of the classes, thereby skewing results away from reflecting true sample membership. To mitigate the introduction of these biases, stemming from the differing class distributions between the two datasets, we employed the forget subset of the test data to ensure a fair and unbiased comparison. Rather than utilizing the entire $\mathcal{D}^t$ for the MIA test, $\mathcal{D}_f$ is combined with $\mathcal{D}_f^t$ to ensure uniform class identification across both subsets, thus mitigating the introduction of class-related biases in $\mathcal{D}^t$. The SVM is trained to discriminate forget train instances vs. forget test instances. The number of forget samples ($N=5000$ for CIFAR10 and $N=500$ for CIFAR100) and forget test samples ($N=1000$ for CIFAR10 and $N=100$ for CIFAR100) are unevenly distributed. To address this imbalance, we resample $\mathcal{D}_f$ to create a less imbalanced $\mathcal{D}_{\text{mia}}$ with a ratio of 1:3 forget test samples per forget samples, resulting in a chance level of $.75$. Results for CR scenario in CIFAR10 and CIFAR100 are reported in Tab. \ref{tab:MIA}

In the HR scenario, both $\mathcal{D}^t$ and $\mathcal{D}_f$ encompass all classes, with an equal number of samples per dataset. Consequently, the chance level for the MIA test in this case is $.5$.

\begin{table}[t]
    \centering
    \input{tables/table_5_supp}
    \caption{Results obtained applying the MIA defined in \cite{Salem2019,song2020,cotogni2023duck} adapted for the CR scenario.}
    \label{tab:MIA}
\end{table}

\section{Baselines}
In this section, we report additional details about the baselines and competitors we included in the comparisons. \\
\textbf{Original}: This is the original model trained on the entire dataset. This model is used as a starting point for SCAR, the competitors, and the baselines considered. The goal of every unlearning algorithm is to remove the knowledge about the forget set from this model. The model, as Resnet18, has been trained for 300 epochs with cosine annealing as $lr$ scheduler.\\
\textbf{Retrained}: The ``Retrained'' model, is a model initialized from random weights and then trained only on the retain set $\mathcal{D}_r$. This is considered an upper bound since it does not have any knowledge of the forget data $\mathcal{D}_f$. The model has been trained for 200 epochs with cosine annealing as $lr$ scheduler.\\
\textbf{Finetuning\cite{golatkar2020eternal}}: This method fine-tunes the original model using the retained data $\mathcal{D}_r$. The model has been fine-tuned for 30 epochs and with X of learning rate. Following \cite{Chen_2023_CVPR} the fine-tuning method results are effective but time-consuming. Importantly, this method requires access to the entire retained set. 
\textbf{Negative Gradient (NG)\cite{golatkar2020eternal}}: In this method, the original model is tuned on the forget data minimizing the inverse of gradient. Results are reported from \cite{cotogni2023duck}. \\
\textbf{Random Label (RL)\cite{hayase2020selective}}: In this method, the original model is tuned, using the cross-entropy loss, matching the exemplars from the forget set with random labels among the ones of the retain data. Results are reported from \cite{cotogni2023duck}. \\
\textbf{Boundary Expanding (BE)\cite{Chen_2023_CVPR}}: This method assigns an additional shadow class to each sample in the forget-set, thereby shifting the decision boundary and leveraging the decision space. \\
\textbf{Boundary Shrink (BS)\cite{Chen_2023_CVPR}}: In this method, the decision boundary is adjusted matching for each sample in the forget-set the closest wrong class label and then finetuning the model with the forget set with these new labels and the retain set using cross-entropy loss. Results are reported from \cite{cotogni2023duck}\\
\textbf{ERM-KTP (ERM)\cite{lin2023erm}}: This approach switches between the Entanglement-Reduced Mask and the Knowledge Transfer and Prohibition phases to erase the data related to the forget-set while enhancing the accuracy on the retain-set. Results are reported from \cite{cotogni2023duck}\\ 
\textbf{SCRUB\cite{kurmanji2023towards}}: This method employs a distillation mechanism to forget the $\mathcal{D}_f$ pushing away the student network predictions (unlearned model) from the teacher network (original model). This procedure is called max-steps. Unfortunately, this mechanism is extremely disruptive, and SCRUB alternates to the max-steps a specific procedure to regain performance on the retain set called min-steps which combines distillation and cross-entropy losses. The distillation loss is weighted .001 whereas the cross-entropy is .999. 
Additionally, the authors conducted a limited number of min-steps to recover the lost knowledge pertaining to the retained data. For each dataset and scenario, lr, number of min-steps, max-steps, and epochs have been optimized to obtain the best overall AUS.\\ 
\textbf{DUCK\cite{cotogni2023duck}}: In this paper, the authors minimize the distance between feature vectors derived from the forget data and the nearest centroid of a different class through metric learning. Concurrently, cross-entropy loss has been employed to maintain performance on retained data. This approach operates directly on single forget samples, enabling the method to function effectively in both CR and HR scenarios. Results are reported from \cite{cotogni2023duck}. \\
\textbf{Bad Teacher\cite{chundawat2023can}}: In this method, similarly to \cite{kurmanji2023towards}, the student model adheres to the original model (i.e., the teacher) with respect to the retain set data. Simultaneously, it minimizes the KL-divergence between its logits and those of a randomly initialized model using as input the forget set data. We followed the training procedure reported in the paper: the model has been trained for 4 epochs with $lr=0.0001$ and Adam as optimizer. \\
\textbf{L1-sparse\cite{jia2023model}}: In this paper the authors elucidates the relationship between exact unlearning methods and approximate unlearning methods when subjected to pruning. This led to the development of an unlearning strategy based on two phases: the pruning phase and the unlearning phase. During the pruning phase, the original model is pruned removing uninformative weights connections. Then, an unlearning regularization, i.e. the ``L1-sparse MU'' method, is applied to the pruned model for 10 epochs with $lr=0.001$. \\

\newpage
\bibliographystyle{splncs04}
\bibliography{arxiv}
\end{document}

%% file: tables/table1.tex
\resizebox{1.\linewidth}{!}{\begin{tabular}{c|cccc|cccc}
\toprule 
Surrogate dataset & \multicolumn{4}{c|}{CIFAR100}                                  & \multicolumn{4}{c}{TinyImagenet} \\
$\mathcal{D}^{\text{sur}}$                  & $\mathcal{A}_r^t$ & $\mathcal{A}_f^t$ & AUS & p-val& $\mathcal{A}_r^t$ & $\mathcal{A}_f^t$ & AUS &p-val \\
\midrule
Random Images & 72.20(01.90) & 02.60(02.17) & 0.922(0.027)& <$10^{-3}$& 60.69(01.80) & 01.20(01.40) & 0.912(0.022)&<$10^{-3}$\\
Imagenet subset & 72.93(01.78) & 02.00(02.00) & \textbf{0.935(0.025)}&<$10^{-3}$& 62.99(01.39)& 00.60(01.37)& \textbf{0.940(0.019)}&<$10^{-3}$\\
COCO subset&71.61(01.92) & 00.84(00.43) & 0.933(0.023)&<$10^{-3}$& 62.04(01.41) & 01.00(01.41) & 0.927(0.019)&<$10^{-3}$\\
Imagenet distilled & 72.13(02.28) & 02.51(01.62) & 0.923(0.023)&<$10^{-3}$ & 62.36(01.56)& 01.03(01.41) & 0.930(0.020)&<$10^{-3}$ \\
Gaussian Noise & 16.62(08.64) & 00.70(00.67) & 0.388(0.086)&<$10^{-3}$& 17.13(06.08) & 01.20(01.93)& 0.482(0.061)&<$10^{-3}$\\

\bottomrule
\end{tabular}}

%% file: tables/table2.tex
\resizebox{1.\linewidth}{!}{\begin{tabular}{l|cc|ccc|ccc|ccc}
\toprule
            & \multirow{2}{*}{\makecell{$\mathcal{D}_r$\\free}}  &  \multirow{2}{*}{\makecell{$\mathcal{D}_f$\\free}}  &  \multicolumn{3}{c|}{CIFAR10}  & \multicolumn{3}{c|}{CIFAR100}  & \multicolumn{3}{c}{TinyImagenet}  \\
            &                          &                           & $\mathcal{A}_r^t$ & $\mathcal{A}_f^t$& AUS & $\mathcal{A}_r^t$ & $\mathcal{A}_f^t$& AUS & $\mathcal{A}_r^t$ & $\mathcal{A}_f^t$& AUS \\
\midrule
Original    &    -                     & -                         & 88.64(00.63) & 88.34(00.62) & 0.531(0.005) & 77.55(00.11) & 77.50(02.80) & 0.563(0.009) & 68.26(00.08) & 64.60(15.64) & 0.607(0.058)\\
Retrained$^\dagger$   & - & - & 88.05(01.28) & 00.00(00.00) & 0.994(0.014) & 77.97(00.42) & 00.00(00.00) & 1.004(0.022) & 67.67(01.00) & 00.00(00.00) & 0.993(0.010)\\
Fine Tuning$^\dagger$ & -  & - & 87.93(01.14) & 00.00(00.00) & 0.993(0.013) & 77.31(02.17) & 00.00(00.00) & 0.998(0.022) & 67.89(00.24) & 00.00(00.00) & 0.994(0.010)\\
\hline
DUCK$^\dagger$        & \textcolor{Red}{\xmark}  & \textcolor{Red}{\xmark}  & 85.53(01.37) & 00.00(00.00) & 0.969(0.015) & 71.57(02.08) & 01.00(01.76) & 0.931(0.026) & 61.29(01.13) & 00.20(00.63) & 0.927(0.013)\\
Boundary S.$^\dagger$ & \textcolor{Red}{\xmark}  & \textcolor{Red}{\xmark}  & 83.81(02.29) & 13.64(03.57) & 0.840(0.543)  & 55.07(11.85) & 03.54(02.26) & 0.749(0.116) & 55.98(03.33) & 04.25(02.12) & 0.840(0.036)\\
Boundary E.$^\dagger$ & \textcolor{Red}{\xmark}  & \textcolor{Red}{\xmark}   & 82.36(02.39) & 13.34(03.21) & 0.827(0.032) & 55.18(11.98) & 03.50(02.32) & 0.750(0.117) & 55.92(03.45) & 02.60(02.50) & 0.853(0.040)\\
SCRUB       & \textcolor{Red}{\xmark} & \textcolor{Red}{\xmark}   & 87.88(01.18)     & 00.00(00.00)     & \cellcolor{red!25}0.992(0.013)     & 77.29(00.25)     & 02.00(05.35)     & \cellcolor{red!25}0.977(0.051)     & 68.15(00.25)           & 01.20(03.79)         & \cellcolor{red!25}0.986(0.037)           \\
L1-Sparse   & \textcolor{Red}{\xmark} & \textcolor{Red}{\xmark}  & 85.21(09.41)     & 04.88(02.18)     & 0.862(0.023)     & 62.63(07.56)     & 01.22(00.53)     & 0.879(0.018)     & 63.76(01.29)         &  04.81(01.54)         &    0.901(0.011)      \\
ERM-KTP$^\dagger$     & \textcolor{Red}{\xmark}  & \textcolor{Red}{\xmark} & 80.31(-)     & 00.00(-)     & 0.917(-)     & 58.96(-)     & 00.00(-)     & 0.814(-)     & 49.45(-)     & 00.00(-)     & 0.811(-)    \\
Bad Teacher& \textcolor{Red}{\xmark}  & \textcolor{Red}{\xmark} &  88.89(00.87) & 02.37(05.86)  & 0.982(0.057)  & 77.19(00.19)  & 06.00(10.09)   & 0.940(0.089)  &  64.09(00.78) & 11.20(09.04)  &   0.862(0.070)  \\
\midrule
Neg. Grad.$^\dagger$  & \textcolor{Green}{\cmark} & \textcolor{Red}{\xmark} & 76.27(03.27) & 00.56(00.12) & 0.871(0.033) & 62.84(06.13) & 00.50(00.50) & 0.849(0.061) & 60.09(02.58) & 00.60(01.35) & 0.911(0.028)\\
Rand. Lab.$^\dagger$  & \textcolor{Green}{\cmark}  & \textcolor{Red}{\xmark} & 65.46(07.59) & 00.83(00.44) & 0.762(0.076) & 55.31(07.06) & 00.40(00.70) & 0.774(0.070) & 43.29(10.10) & 01.20(01.03) & 0.740(0.100)\\
\textbf{SCAR}         & \textcolor{Green}{\cmark} & \textcolor{Red}{\xmark}  &     87.71(01.62)         &  00.97(00.24)            &      \cellcolor{green!25}0.981(0.017)        & 72.93(01.78) & 02.00(02.00) & \cellcolor{green!25}0.935(0.025)&  62.99(01.39)& 00.60(01.37)&\cellcolor{green!25}0.940(0.019)\\
\textbf{SCAR Self-Forget}         & \textcolor{Green}{\cmark}& \textcolor{Green}{\cmark} &       87.18(01.76)       &     01.20(00.43)         &       0.974(0.019)       &       71.09(02.69)       &   00.70(00.95)           &    0.929(0.028)          &       60.79(02.15)     &      00.80(01.40)        &       0.917(0.025)    \\
\bottomrule
\end{tabular}}

%% file: tables/table3.tex
\resizebox{1.\linewidth}{!}{\begin{tabular}{l|c|cccc|cccc|cccc}
\toprule
                &    \multirow{2}{*}{\makecell{ $\mathcal{D}_r$\\free }}        & \multicolumn{4}{c|}{CIFAR10}                                                 & \multicolumn{4}{c|}{CIFAR100}                                                & \multicolumn{4}{c}{TinyImagenet}                                                    \\
                &  & $\mathcal{A}^t$ & $\mathcal{A}_f$ &  $\mathcal{F}_1$& AUS & $\mathcal{A}^t$ &$\mathcal{A}_f$ &  $\mathcal{F}_1$& AUS & $\mathcal{A}^t$ & $\mathcal{A}_f$ &  $\mathcal{F}_1$& AUS\\
\midrule
Original &    -     & 88.54(00.25) & 99.49(00.08) &  54.62(00.80) & 0.900(0.004) & 77.23(00.53) &99.63(00.10) &  61.80(00.59) & 0.814(0.006) &68.08(00.39) & 84.83(00.44) &  54.06(01.48) & 0.854(0.007) \\

Retrained$^\dagger$ &     - & 84.13(00.59) &84.56(00.72) &  50.22(00.80) & 0.950(0.011) & 77.27(01.03) &76.87(00.95) &  49.45(00.40) & 0.993(0.017) &63.45(02.34) & 63.24(02.58) &  49.56(00.47) & 0.950(0.040)\\

Fine Tuning$^\dagger$ &    -  & 85.70(00.48) &88.66(00.44) &  50.06(01.30) & 0.942(0.008) & 72.06(00.52) &74.97(00.72) &  49.68(00.43) & 0.918(0.009) & 67.03(00.45) &70.52(00.65) &  49.86(00.56) & 0.937(0.011)\\
\hline

DUCK$^\dagger$ &      \textcolor{Red}{\xmark} & 85.96(00.64) &86.05(00.55) &  50.26(00.49) & \cellcolor{red!25}0.972(0.011) & 74.74(00.74) &75.51(02.01) &  50.66(00.64) & \cellcolor{red!25}0.965(0.022) & 62.01(00.79) &64.43(00.76) &  49.87(00.73) & \cellcolor{red!25}0.916(0.014) \\

SCRUB &      \textcolor{Red}{\xmark} & 78.02(07.80) &82.53(10.80) &  50.59(00.89) & 0.854(0.132) & 73.33(00.85) &92.46(01.50) &  56.45(01.48) & 0.804(0.014) & 68.52(00.40) &83.14(00.53) &  53.89(00.96) & 0.875(0.008)\\

L1-Sparse & \textcolor{Red}{\xmark} & 93.05(00.53) &99.06(00.65) &  52.21(00.95) & 0.927(0.011) & 52.26(01.65) &66.89(02.34) &  53.46(00.94) & 0.690(0.024) & 63.76(01.19) &84.30(01.04) &  53.29(00.73) &0.794(0.032)\\
Bad Teacher & \textcolor{Red}{\xmark} & 86.66(00.44) & 92.91(00.57) &  50.26(00.87)& 0.922(0.008) & 73.57(00.61) & 86.57(00.57)  &  54.44(01.81)  & 0.851(0.006) & 67.41(00.49) & 80.28(00.81) &  54.83(01.44)  & 0.876(0.006)\\
\midrule
Neg. Grad. $^\dagger$ &      \textcolor{Green}{\cmark} & 79.35(00.85) & 87.11(00.86) &  51.74(00.54) & 0.841(0.012) & 60.83(00.77) & 76.77(00.57) &  50.98(00.47) &  0.718(0.009) & 54.83(00.81) & 67.27(00.59) &  49.78(00.33) & 0.770(0.011) \\
Rand. Lab. $^\dagger$&      \textcolor{Green}{\cmark} & 77.28(01.37) & 87.07(01.20) &  51.45(00.79) & 0.807(0.018) & 60.51(00.82) & 77.18(00.50) &  50.55(00.49) & 0.711(0.009) & 54.34(00.85) & 67.65(00.45) &  49.84(00.64) & 0.760(0.011) \\
\textbf{SCAR} &      \textcolor{Green}{\cmark} &86.14(00.47) &88.40(00.31) &50.07(00.59) &\cellcolor{green!25}0.953(0.008) &73.23(00.74) & 75.63(00.54) &49.77(00.28) &\cellcolor{green!25}0.934(0.011) & 61.35(00.69)& 66.51(00.44) &  49.85(00.19)& \cellcolor{green!25}0.886(0.011)\\
\bottomrule
\end{tabular}}

%% file: tables/table4.tex
\resizebox{.9\linewidth}{!}{\begin{tabular}{ccc|ccc|ccc}
\toprule
&\multirow{2}{*}{$\mathcal{L}_\text{TD}$}&\multirow{2}{*}{$\mathcal{L}_\text{M}$}& \multicolumn{3}{c|}{CIFAR100}& \multicolumn{3}{c}{TinyImagenet} \\
&& & $\mathcal{A}_r^t$ & $\mathcal{A}_f^t$ & AUS & $\mathcal{A}_r^t$ & $\mathcal{A}_f^t$ & AUS \\
\midrule
\multirow{4}{*}{CR}&\xmark & \xmark &  77.55(00.11) & 77.50(02.80) & 0.563(0.009) & 68.26(00.08) & 64.60(15.64) & 0.607(0.058) \\
&\cmark & \xmark  & 72.09(02.72) & 40.00(13.36)& 0.675(0.067)& 63.14(01.77)& 29.80(12.09)&0.730(0.069) \\
&\xmark & \cmark & 66.90(05.03) & 02.60(02.27) & 0.871(0.053) &49.72(02.54)& 00.80(01.40)& 0.807(0.028)\\
&\cmark & \cmark & 72.93(01.78) & 02.00(02.00) & \textbf{0.935(0.025)} & 62.99(01.39)& 00.60(01.37)&\textbf{0.940(0.019)} \\
\midrule
&& &  $\mathcal{A}^t$ & $\mathcal{A}_f$  & AUS & $\mathcal{A}^t$ & $\mathcal{A}_f$ &  AUS\\
\midrule
\multirow{4}{*}{HR}&\xmark & \xmark &  77.39(00.46) & 99.64(00.09) &  0.817(0.005) & 68.08(00.39) & 84.83(00.44) &  0.854(0.007) \\

&\cmark & \xmark  &76.53(00.41) &99.18(00.14) &  0.807(0.005) &64.84(00.60)&80.51(00.51)  & 0.835(0.009) \\

&\xmark & \cmark & 18.83(01.06)&19.69(01.19) &  0.409(0.013)&  05.88(02.94) &05.51(02.63) & 0.376(0.033) \\
&\cmark & \cmark &73.23(00.74) &75.63(00.54) &\textbf{0.934(0.011)}& 61.35(00.69)& 66.51(00.44) &  \textbf{0.886(0.011)}\\
\bottomrule
\end{tabular}}




%% file: tables/table5.tex
\resizebox{.9\linewidth}{!}{\begin{tabular}{cl|ccc|ccc}
\toprule
& & \multicolumn{3}{c|}{CIFAR100}& \multicolumn{3}{c}{TinyImagenet} \\

& & $\mathcal{A}_r^t$ & $\mathcal{A}_f^t$ & AUS & $\mathcal{A}_r^t$ & $\mathcal{A}_f^t$ & AUS \\
\midrule
\multirow{4}{*}{CR}&Cosine Similarity & 72.74(02.19)& 02.00(02.36) & 0.933(0.031)& 60.55(01.66) & 01.00(01.41)&0.912(0.021) \\

&$\mathcal{L}_2$ Distance & 72.45(02.23) & 02.70(02.54) & 0.924(0.032) & 57.74(01.66)& 01.40(01.90)&0.881(0.023) \\

&Mahalanobis & 72.93(01.78) & 02.00(02.00) & \textbf{0.935(0.025)}& 62.99(01.39)& 00.60(01.37)&\textbf{0.940(0.019)}\\
\midrule
& & $\mathcal{A}^t$ & $\mathcal{A}_f$ & AUS & $\mathcal{A}^t$ & $\mathcal{A}_f$ & AUS \\
\midrule
\multirow{4}{*}{HR}& Cosine Similarity & 73.88(00.46) & 77.05(00.54) & 0.934(0.008) &61.27(00.81) &66.48(00.52) & 0.883(0.012) \\

&$\mathcal{L}_2$ Distance & 69.58(00.82) & 76.93(00.55) & 0.857(0.011) & 49.20(01.18) &48.69(01.01) & 0.806(0.018) \\

&Mahalanobis &73.23(00.74)& 75.63(00.54) &\textbf{0.934(0.011)} & 61.35(00.69)& 66.51(00.44) & \textbf{0.886(0.011)}\\
\bottomrule
\end{tabular}}





%% file: tables/table6.tex
\resizebox{1.\linewidth}{!}{\begin{tabular}{l|ccc}
\toprule
& \multicolumn{3}{c}{CIFAR100}  \\                                     
& $\mathcal{A}_r^t$ & $\mathcal{A}_f^t$ & AUS \\
\midrule
Original (AllCNN) & 70.08(00.15)  & 69.70(14.83) & 0.545(0.048) \\
Unlearned &65.35(02.37) & 02.90(03.38) & 0.926(0.036)  \\
\midrule
Original (Resnet18) &  77.55(00.11) & 77.50(02.80) & 0.563(0.009) \\ 
Unlearned &72.93(01.78) & 02.00(02.00) & 0.935(0.025)\\
\midrule
Original (Resnet34) & 80.13(00.10) & 80.30(10.02) & 0.569(0.031)  \\
Unlearned &75.67(01.74) & 01.90(00.92) & 0.937(0.019)\\
\midrule
Original (Resnet50) & 80.64(00.10) & 80.80(09.50) &0.570(0.030)\\
Unlearned & 75.99(01.77) & 00.80(02.36) & 0.942(0.028) \\
\midrule
Original (ViT-B16) & 85.51(01.45) &  83.30(06.24)  & 0.546(0.023) \\
Unlearned & 81.99(02.59) & 02.04(03.10) & 0.945(0.031)  \\
\bottomrule
\end{tabular}}

%% file: tables/table_1_suppl.tex
\resizebox{1.\linewidth}{!}{\begin{tabular}{cr|ccc|ccc}
\toprule
& & \multicolumn{3}{c|}{CIFAR100}& \multicolumn{3}{c}{TinyImagenet} \\

& & $\mathcal{A}_r^t$ & $\mathcal{A}_f^t$ & AUS & $\mathcal{A}_r^t$ & $\mathcal{A}_f^t$ & AUS \\
\midrule
\multirow{4}{*}{\rotatebox{90}{CR}}&SCAR (Imagenet) &72.93(01.78) & 02.00(02.00) & 0.935(0.025)&  62.99(01.39)& 00.60(01.37)&0.940(0.019)\\
&SCAR S-F(Imagenet)  & 71.69(02.69) &   00.70(00.95)           &    0.929(0.028)          &       60.79(02.15)     &      00.80(01.40)        &       0.917(0.025)    \\
&SCAR (COCO) & 71.61(01.92) & 00.84(00.43) & 0.933(0.023)& 62.04(01.41) & 01.00(01.41) & 0.927(0.019)\\
&SCAR S-F(COCO) & 71.27(02.95) & 00.75(01.30) & 0.930(0.032)&61.88(00.98)& 01.20(01.69)&0.924(0.018)\\
\midrule
& & $\mathcal{A}^t$ & $\mathcal{A}_f$ & AUS & $\mathcal{A}^t$ & $\mathcal{A}_f$ & AUS \\
\midrule
\multirow{2}{*}{\rotatebox{90}{HR}}&SCAR (Imagenet) &73.23(00.74)& 75.63(00.54) &0.934(0.011) & 61.35(00.69)& 66.51(00.44) & 0.886(0.011)\\
&SCAR (COCO) &72.51(00.41)& 75.64(00.57) &0.921(0.008) & 61.56(00.73)& 66.49(00.47) & 0.890(0.011)\\

\bottomrule
\end{tabular}}

%% file: tables/table_2_supp.tex
\resizebox{1.\linewidth}{!}{\begin{tabular}{lccccccc}
\toprule
Method & \makecell{$\mathcal{D}_r$\\free}& \makecell{$\mathcal{D}_f$\\free} & AUS & Time(s) & \makecell{Retain-Data \\Stored(Mb)} & \makecell{Forget-Data\\ Stored(Mb)} & \makecell{Total-Data\\Stored(Mb)}\\
\midrule
Fine Tuning & \textcolor{Red}{\xmark} & \textcolor{Green}{\cmark}& 0.998(0.022) & 366.99 & 145.02 &  0 &145.02\\
DUCK &\textcolor{Red}{\xmark} & \textcolor{Red}{\xmark} & 0.931(0.026) & 12.56& 145.02 & 1.46&146.48\\
Boundary S. &\textcolor{Red}{\xmark}& \textcolor{Red}{\xmark} & 0.749(0.116)& 23.19 &145.02 & 1.46&146.48 \\
Boundary E. &\textcolor{Red}{\xmark}& \textcolor{Red}{\xmark} & 0.750(0.117)&  66.50 &145.02&1.46& 146.48\\
SCRUB &\textcolor{Red}{\xmark}& \textcolor{Red}{\xmark}& 0.977(0.051) &139.18 &145.02 &1.46&146.48 \\
L1-Sparse &\textcolor{Red}{\xmark} &\textcolor{Red}{\xmark}& 0.879(0.018)&35.29 & 145.02&1.46&146.48\\
ERM-KTP &\textcolor{Red}{\xmark}& \textcolor{Red}{\xmark}& 0.814(-)& 550.14&145.02 &1.46&146.48\\
Bad Teacher &\textcolor{Red}{\xmark} &\textcolor{Red}{\xmark}&0.940(0.089) & 56.07 &145.02 &1.46&146.48\\
\midrule
Neg. Grad. &\textcolor{Green}{\cmark} &\textcolor{Red}{\xmark}& 0.849(0.061) &09.02 & 0 &1.46&1.46\\
Rand. Lab.&\textcolor{Green}{\cmark} & \textcolor{Red}{\xmark}& 0.774(0.070)&17.03 & 0&1.46&1.46\\
\textbf{SCAR} & \textcolor{Green}{\cmark}&\textcolor{Red}{\xmark}& 0.935(0.025)&109.11 & 29.30 &1.46&30.76\\
\textbf{SCAR Self-Forget} &\textcolor{Green}{\cmark} &\textcolor{Green}{\cmark}& 0.929(0.028)&101.61 & 29.30 &0&29.30\\
\bottomrule
\end{tabular}}

%% file: tables/table_3_supp.tex
\resizebox{1.\linewidth}{!}{\begin{tabular}{lccccccccc|ccccccccc}
\toprule
&\multicolumn{9}{c|}{CR}&\multicolumn{9}{c}{HR}\\
            &$\lambda_1$&$\lambda_2$&$lr$&BS&$T$&$\delta$&$\gamma_1$&$\gamma_2$&E$_{\text{max}}$ &$\lambda_1$&$\lambda_2$&$lr$&BS&$T$&$\delta$&$\gamma_1$&$\gamma_2$&E$_{\text{max}}$\\
\midrule
Cifar10     &1&5&$5\times 10^{-4}$&1024&1&0.5&3&3&30
            &1&8&$5\times 10^{-4}$&1024&5&1&3&3&30\\ 
Cifar100    &1&5&$5\times 10^{-4}$&1024&1&0.5&3&3&30 
            &1&6&$5\times 10^{-4}$&1024&2.5&1&3&3&30\\ 
TinyImagenet&1&5&$1\times 10^{-4}$&1024&1&0.5&3&3&30 
            &1&5&$1\times 10^{-4}$&1024&1.8&1&6&6&30\\ 
\midrule

\end{tabular}}

%% file: tables/table_4_supp.tex
\begin{tabular}{lccccccccc}

\toprule
            &$\lambda_1$&$\lambda_2$&$lr$&BS&$T$&$\delta$&$\gamma_1$&$\gamma_2$&Epoch$_{\text{max}}$\\
\midrule
Cifar10     &1&5&$7.5\times 10^{-4}$&1024&2&0.5&3&3&25\\
Cifar100    &1&4&$1\times 10^{-3}$&1024&2&0.5&3&3&25\\ 
           
TinyImagenet&1&5&$1\times 10^{-4}$&1024&2&0.5&3&3&25\\ 
            
\midrule

\end{tabular}

%% file: tables/table_5_supp.tex

\resizebox{1.\linewidth}{!}{\begin{tabular}{l|cccc|cccc}
\toprule
                & \multicolumn{4}{c|}{CIFAR10}                                 & \multicolumn{4}{c}{CIFAR100}\\
          & $\mathcal{A}^t$ & $\mathcal{A}_f$ & AUS&  $\mathcal{F}_1$ & $\mathcal{A}^t$ &$\mathcal{A}_f$ &  AUS &$\mathcal{F}_1$\\
\midrule
Original   & 88.64(00.63) & 88.34(00.62) & 0.531(0.005)&77.63(1.33) & 77.55(00.11) & 77.50(02.80) & 0.563(0.009)&83.15(4.12)\\
Retrained& 88.05(01.28) & 00.00(00.00) & 0.994(0.014)&74.91(0.10) & 77.97(00.42) & 00.00(00.00) & 1.004(0.022)&75.47(0.75)\\
\textbf{SCAR} &     87.71(01.62)         &  00.97(00.24)            &     0.981(0.017)&74.91(0.19)        & 72.93(01.78) & 02.00(02.00) & 0.935(0.025)&75.60(0.44)\\
\bottomrule
\end{tabular}}

%% file: arxiv.bbl
\begin{thebibliography}{10}
\providecommand{\url}[1]{\texttt{#1}}
\providecommand{\urlprefix}{URL }
\providecommand{\doi}[1]{https://doi.org/#1}

\bibitem{randomIMG}
Image captioning dataset, random images. \url{https://www.kaggle.com/datasets/shamsaddin97/image-captioning-dataset-random-images/data}, accessed: 2024-02-14

\bibitem{bibas2021single}
Bibas, K., Feder, M., Hassner, T.: Single layer predictive normalized maximum likelihood for out-of-distribution detection. Advances in Neural Information Processing Systems  \textbf{34},  1179--1191 (2021)

\bibitem{bourtoule2021machine}
Bourtoule, L., Chandrasekaran, V., Choquette-Choo, C.A., Jia, H., Travers, A., Zhang, B., Lie, D., Papernot, N.: Machine unlearning. In: 2021 IEEE Symposium on Security and Privacy (SP). pp. 141--159. IEEE (2021)

\bibitem{Chen_2023_CVPR}
Chen, M., Gao, W., Liu, G., Peng, K., Wang, C.: Boundary unlearning: Rapid forgetting of deep networks via shifting the decision boundary. In: Proceedings of the IEEE/CVF Conference on Computer Vision and Pattern Recognition (CVPR). pp. 7766--7775 (June 2023)

\bibitem{chen2021MI}
Chen, S., Kahla, M., Jia, R., Qi, G.J.: Knowledge-enriched distributional model inversion attacks pp. 16158--16167 (2021). \doi{10.1109/ICCV48922.2021.01587}

\bibitem{Chen2009Shrinkage}
Chen, Y., Wiesel, A., Eldar, Y.C., Hero, A.O.: Shrinkage algorithms for mmse covariance estimation. IEEE Transactions on Signal Processing  \textbf{58},  5016--5029 (2009), \url{https://api.semanticscholar.org/CorpusID:12616083}

\bibitem{cheng2023multimodal}
Cheng, J., Amiri, H.: Multimodal machine unlearning. arXiv preprint arXiv:2311.12047  (2023)

\bibitem{chundawat2023can}
Chundawat, V.S., Tarun, A.K., Mandal, M., Kankanhalli, M.: Can bad teaching induce forgetting? unlearning in deep networks using an incompetent teacher. In: Proceedings of the AAAI Conference on Artificial Intelligence. vol.~37, pp. 7210--7217 (2023)

\bibitem{cotogni2023duck}
Cotogni, M., Bonato, J., Sabetta, L., Pelosin, F., Nicolosi, A.: Duck: Distance-based unlearning via centroid kinematics. arXiv preprint arXiv:2312.02052  (2023)

\bibitem{Lange2021_proto}
De~Lange, M., Tuytelaars, T.: Continual prototype evolution: Learning online from non-stationary data streams (2021). \doi{10.1109/ICCV48922.2021.00814}

\bibitem{deng2009imagenet}
Deng, J., Dong, W., Socher, R., Li, L.J., Li, K., Fei-Fei, L.: Imagenet: A large-scale hierarchical image database. In: 2009 IEEE conference on computer vision and pattern recognition. pp. 248--255. Ieee (2009)

\bibitem{Du2021_fairness}
Du, M., Yang, F., Zou, N., Hu, X.: Fairness in deep learning: A computational perspective. IEEE Intelligent Systems  \textbf{36}(4),  25--34 (2021). \doi{10.1109/MIS.2020.3000681}

\bibitem{gongfag2021_KDOOD}
Fang, G., Bao, Y., Song, J., Wang, X., Xie, D., Shen, C., Song, M.: Mosaicking to distill: Knowledge distillation from out-of-domain data. In: Ranzato, M., Beygelzimer, A., Dauphin, Y., Liang, P., Vaughan, J.W. (eds.) Advances in Neural Information Processing Systems. vol.~34, pp. 11920--11932. Curran Associates, Inc. (2021), \url{https://proceedings.neurips.cc/paper_files/paper/2021/file/63dc7ed1010d3c3b8269faf0ba7491d4-Paper.pdf}

\bibitem{felps2020class}
Felps, D.L., Schwickerath, A.D., Williams, J.D., Vuong, T.N., Briggs, A., Hunt, M., Sakmar, E., Saranchak, D.D., Shumaker, T.: Class clown: Data redaction in machine unlearning at enterprise scale. arXiv preprint arXiv:2012.04699  (2020)

\bibitem{fort2021exploring}
Fort, S., Ren, J., Lakshminarayanan, B.: Exploring the limits of out-of-distribution detection. Advances in Neural Information Processing Systems  \textbf{34},  7068--7081 (2021)

\bibitem{Fredrikson2015MI}
Fredrikson, M., Jha, S., Ristenpart, T.: Model inversion attacks that exploit confidence information and basic countermeasures. Proceedings of the 22nd ACM SIGSAC Conference on Computer and Communications Security  (2015), \url{https://api.semanticscholar.org/CorpusID:207229839}

\bibitem{golatkar2020eternal}
Golatkar, A., Achille, A., Soatto, S.: Eternal sunshine of the spotless net: Selective forgetting in deep networks. In: Proceedings of the IEEE/CVF Conference on Computer Vision and Pattern Recognition. pp. 9304--9312 (2020)

\bibitem{golatkar2020forgetting}
Golatkar, A., Achille, A., Soatto, S.: Forgetting outside the box: Scrubbing deep networks of information accessible from input-output observations. In: Computer Vision--ECCV 2020: 16th European Conference, Glasgow, UK, August 23--28, 2020, Proceedings, Part XXIX 16. pp. 383--398. Springer (2020)

\bibitem{graves2021amnesiac}
Graves, L., Nagisetty, V., Ganesh, V.: Amnesiac machine learning. In: Proceedings of the AAAI Conference on Artificial Intelligence. vol.~35, pp. 11516--11524 (2021)

\bibitem{pmlr-v119-guo20c}
Guo, C., Goldstein, T., Hannun, A., Van Der~Maaten, L.: Certified data removal from machine learning models. In: Proceedings of the 37th International Conference on Machine Learning. Proceedings of Machine Learning Research, vol.~119, pp. 3832--3842. PMLR (2020)

\bibitem{guo2017calibration}
Guo, C., Pleiss, G., Sun, Y., Weinberger, K.Q.: On calibration of modern neural networks. In: International conference on machine learning. pp. 1321--1330. PMLR (2017)

\bibitem{xu2020_proto}
Han, X., Dai, Y., Gao, T., Lin, Y., Liu, Z., Li, P., Sun, M., Zhou, J.: Continual relation learning via episodic memory activation and reconsolidation (2020). \doi{10.18653/v1/2020.acl-main.573}, \url{https://aclanthology.org/2020.acl-main.573}

\bibitem{hayase2020selective}
Hayase, T., Yasutomi, S., Katoh, T.: Selective forgetting of deep networks at a finer level than samples. arXiv preprint arXiv:2012.11849  (2020)

\bibitem{he2016deep}
He, K., Zhang, X., Ren, S., Sun, J.: Deep residual learning for image recognition. In: Proceedings of the IEEE conference on computer vision and pattern recognition. pp. 770--778 (2016)

\bibitem{hendrycks2017baseline}
Hendrycks, D., Gimpel, K.: A baseline for detecting misclassified and out-of-distribution examples in neural networks. In: International Conference on Learning Representations (2017), \url{https://openreview.net/forum?id=Hkg4TI9xl}

\bibitem{hinton2015distilling}
Hinton, G., Vinyals, O., Dean, J.: Distilling the knowledge in a neural network. arXiv preprint arXiv:1503.02531  (2015)

\bibitem{hsu2020generalized}
Hsu, Y.C., Shen, Y., Jin, H., Kira, Z.: Generalized odin: Detecting out-of-distribution image without learning from out-of-distribution data. In: Proceedings of the IEEE/CVF Conference on Computer Vision and Pattern Recognition. pp. 10951--10960 (2020)

\bibitem{jia2023model}
Jia, J., Liu, J., Ram, P., Yao, Y., Liu, G., Liu, Y., Sharma, P., Liu, S.: Model sparsity can simplify machine unlearning. In: Annual Conference on Neural Information Processing Systems (2023)

\bibitem{Kahla2022MI}
Kahla, M., Chen, S., Just, H.A., Jia, R.: Label-only model inversion attacks via boundary repulsion. 2022 IEEE/CVF Conference on Computer Vision and Pattern Recognition (CVPR) pp. 15025--15033 (2022). \doi{10.1109/CVPR52688.2022.01462}

\bibitem{kim2022efficient}
Kim, J., Woo, S.S.: Efficient two-stage model retraining for machine unlearning. In: Proceedings of the IEEE/CVF Conference on Computer Vision and Pattern Recognition. pp. 4361--4369 (2022)

\bibitem{krizhevsky2009learning}
Krizhevsky, A., Hinton, G., et~al.: Learning multiple layers of features from tiny images  (2009)

\bibitem{kurmanji2023towards}
Kurmanji, M., Triantafillou, P., Hayes, J., Triantafillou, E.: Towards unbounded machine unlearning. In: Thirty-seventh Conference on Neural Information Processing Systems (2023), \url{https://openreview.net/forum?id=OveBaTtUAT}

\bibitem{le2015tiny}
Le, Y., Yang, X.: Tiny imagenet visual recognition challenge. CS 231N  \textbf{7}(7), ~3 (2015)

\bibitem{leibig2017leveraging}
Leibig, C., Allken, V., Ayhan, M.S., Berens, P., Wahl, S.: Leveraging uncertainty information from deep neural networks for disease detection. Scientific reports  \textbf{7}(1),  17816 (2017)

\bibitem{lin2023erm}
Lin, S., Zhang, X., Chen, C., Chen, X., Susilo, W.: Erm-ktp: Knowledge-level machine unlearning via knowledge transfer. In: Proceedings of the IEEE/CVF Conference on Computer Vision and Pattern Recognition. pp. 20147--20155 (2023)

\bibitem{lin2014microsoft}
Lin, T.Y., Maire, M., Belongie, S., Hays, J., Perona, P., Ramanan, D., Doll{\'a}r, P., Zitnick, C.L.: Microsoft coco: Common objects in context. In: Computer Vision--ECCV 2014: 13th European Conference, Zurich, Switzerland, September 6-12, 2014, Proceedings, Part V 13. pp. 740--755. Springer (2014)

\bibitem{Liu_2023_ICCV}
Liu, J., Xue, M., Lou, J., Zhang, X., Xiong, L., Qin, Z.: Muter: Machine unlearning on adversarially trained models. In: Proceedings of the IEEE/CVF International Conference on Computer Vision (ICCV). pp. 4892--4902 (October 2023)

\bibitem{Liu2021PrivacyAS}
Liu, X., Xie, L., Wang, Y., Zou, J., Xiong, J., Ying, Z., Vasilakos, A.V.: Privacy and security issues in deep learning: A survey. IEEE Access  \textbf{9},  4566--4593 (2021), \url{https://api.semanticscholar.org/CorpusID:231423554}

\bibitem{loshchilov2017decoupled}
Loshchilov, I., Hutter, F.: Decoupled weight decay regularization. arXiv preprint arXiv:1711.05101  (2017)

\bibitem{van2008visualizing}
Van~der Maaten, L., Hinton, G.: Visualizing data using t-sne. Journal of Machine Learning Research  \textbf{9}(11) (2008)

\bibitem{magdziarczyk2019right}
Magdziarczyk, M.: Right to be forgotten in light of regulation (eu) 2016/679 of the european parliament and of the council of 27 april 2016 on the protection of natural persons with regard to the processing of personal data and on the free movement of such data, and repealing directive 95/46/ec. In: 6th International Multidisciplinary Scientific Conference on Social Sciences and Art Sgem 2019. pp. 177--184 (2019)

\bibitem{massey1951kolmogorov}
Massey~Jr, F.J.: The kolmogorov-smirnov test for goodness of fit. Journal of the American statistical Association  \textbf{46}(253),  68--78 (1951)

\bibitem{mercuri2022introduction}
Mercuri, S., Khraishi, R., Okhrati, R., Batra, D., Hamill, C., Ghasempour, T., Nowlan, A.: An introduction to machine unlearning. arXiv preprint arXiv:2209.00939  (2022)

\bibitem{neel2021descent}
Neel, S., Roth, A., Sharifi-Malvajerdi, S.: Descent-to-delete: Gradient-based methods for machine unlearning. In: Algorithmic Learning Theory. pp. 931--962. PMLR (2021)

\bibitem{nguyen2015deep}
Nguyen, A., Yosinski, J., Clune, J.: Deep neural networks are easily fooled: High confidence predictions for unrecognizable images. In: Proceedings of the IEEE conference on computer vision and pattern recognition. pp. 427--436 (2015)

\bibitem{nguyen2022survey}
Nguyen, T.T., Huynh, T.T., Nguyen, P.L., Liew, A.W.C., Yin, H., Nguyen, Q.V.H.: A survey of machine unlearning. arXiv preprint arXiv:2209.02299  (2022)

\bibitem{pardau2018california}
Pardau, S.L.: The california consumer privacy act: Towards a european-style privacy regime in the united states. J. Tech. L. \& Pol'y  \textbf{23}, ~68 (2018)

\bibitem{poppi2024removing}
Poppi, S., Poppi, T., Cocchi, F., Cornia, M., Baraldi, L., Cucchiara, R.: {Removing NSFW Concepts from Vision-and-Language Models for Text-to-Image Retrieval and Generation}. arXiv preprint arXiv:2311.16254  (2023)

\bibitem{rebuffi2017_proto}
Rebuffi, S.A., Kolesnikov, A., Sperl, G., Lampert, C.H.: icarl: Incremental classifier and representation learning (2017). \doi{10.1109/CVPR.2017.587}

\bibitem{Rigaki2023}
Rigaki, M., Garcia, S.: A survey of privacy attacks in machine learning. ACM Computing Surveys  \textbf{56}(4),  1–34 (Nov 2023). \doi{10.1145/3624010}, \url{http://dx.doi.org/10.1145/3624010}

\bibitem{Salem2019}
Salem, A., Zhang, Y., Humbert, M., Berrang, P., Fritz, M., Backes, M.: Ml-leaks: Model and data independent membership inference attacks and defenses on machine learning models. In: Proceedings of the 26th Annual Network and Distributed System Security Symposium (NDSS) (2019)

\bibitem{setlur2022adversarial}
Setlur, A., Eysenbach, B., Smith, V., Levine, S.: Adversarial unlearning: Reducing confidence along adversarial directions. Advances in Neural Information Processing Systems  \textbf{35},  18556--18570 (2022)

\bibitem{shaik2023exploring}
Shaik, T.B., Tao, X., Xie, H., Li, L., Zhu, X., Li, Q.: Exploring the landscape of machine unlearning: A comprehensive survey and taxonomy. CoRR  (2023)

\bibitem{Shokri2017}
Shokri, R., Stronati, M., Song, C., Shmatikov, V.: Membership inference attacks against machine learning models. 2017 IEEE Symposium on Security and Privacy (SP) pp. 3--18 (2017)

\bibitem{sommer2020towards}
Sommer, D.M., Song, L., Wagh, S., Mittal, P.: Towards probabilistic verification of machine unlearning. arXiv preprint arXiv:2003.04247  (2020)

\bibitem{song2020}
Song, L., Mittal, P.: Systematic evaluation of privacy risks of machine learning models (2020)

\bibitem{Song2019}
Song, L., Shokri, R., Mittal, P.: Privacy risks of securing machine learning models against adversarial examples. Proceedings of the 2019 ACM SIGSAC Conference on Computer and Communications Security  (2019), \url{https://api.semanticscholar.org/CorpusID:165163934}

\bibitem{tarun2023fast}
Tarun, A.K., Chundawat, V.S., Mandal, M., Kankanhalli, M.: Fast yet effective machine unlearning. IEEE Transactions on Neural Networks and Learning Systems  (2023)

\bibitem{tiwary2023adapt}
Tiwary, P., Guha, A., Panda, S., et~al.: Adapt then unlearn: Exploiting parameter space semantics for unlearning in generative adversarial networks. arXiv preprint arXiv:2309.14054  (2023)

\bibitem{ttukey77}
Tukey, J.W.: Exploratory Data Analysis. Addison-Wesley (1977)

\bibitem{wang2021MI}
Wang, K.C., FU, Y., Li, K., Khisti, A., Zemel, R., Makhzani, A.: Variational model inversion attacks. Advances in Neural Information Processing Systems  \textbf{34},  9706--9719 (2021), \url{https://proceedings.neurips.cc/paper_files/paper/2021/file/50a074e6a8da4662ae0a29edde722179-Paper.pdf}

\bibitem{wu2022puma}
Wu, G., Hashemi, M., Srinivasa, C.: Puma: Performance unchanged model augmentation for training data removal. In: Proceedings of the AAAI Conference on Artificial Intelligence. vol.~36, pp. 8675--8682 (2022)

\bibitem{wu2020deltagrad}
Wu, Y., Dobriban, E., Davidson, S.: Deltagrad: Rapid retraining of machine learning models. In: International Conference on Machine Learning. pp. 10355--10366. PMLR (2020)

\bibitem{xu2023machine}
Xu, J., Wu, Z., Wang, C., Jia, X.: Machine unlearning: Solutions and challenges. arXiv preprint arXiv:2308.07061  (2023)

\bibitem{Yang2019MI}
Yang, Z., Zhang, J., Chang, E.C., Liang, Z.: Neural network inversion in adversarial setting via background knowledge alignment. Proceedings of the 2019 ACM SIGSAC Conference on Computer and Communications Security p. 225–240 (2019). \doi{10.1145/3319535.3354261}, \url{https://doi.org/10.1145/3319535.3354261}

\bibitem{Yeom2017}
Yeom, S., Giacomelli, I., Fredrikson, M., Jha, S.: Privacy risk in machine learning: Analyzing the connection to overfitting. 2018 IEEE 31st Computer Security Foundations Symposium (CSF) pp. 268--282 (2017), \url{https://api.semanticscholar.org/CorpusID:2656445}

\bibitem{yin2024squeeze}
Yin, Z., Xing, E., Shen, Z.: Squeeze, recover and relabel: Dataset condensation at imagenet scale from a new perspective. Advances in Neural Information Processing Systems  \textbf{36} (2024)

\bibitem{zhang2020MI}
Zhang, Y., Jia, R., Pei, H., Wang, W., Li, B., Song, D.: The secret revealer: Generative model-inversion attacks against deep neural networks. 2020 IEEE/CVF Conference on Computer Vision and Pattern Recognition (CVPR) pp. 250--258 (jun 2020). \doi{10.1109/CVPR42600.2020.00033}, \url{https://doi.ieeecomputersociety.org/10.1109/CVPR42600.2020.00033}

\bibitem{zhao2021MI}
Zhao, X., Zhang, W., Xiao, X., Lim, B.: Exploiting explanations for model inversion attacks. 2021 IEEE/CVF International Conference on Computer Vision (ICCV) pp. 662--672 (oct 2021). \doi{10.1109/ICCV48922.2021.00072}, \url{https://doi.ieeecomputersociety.org/10.1109/ICCV48922.2021.00072}

\end{thebibliography}
